\pdfoutput=1
\PassOptionsToPackage{dvipsnames}{xcolor}
\PassOptionsToPackage{hyphens}{url}
\documentclass{MITcsail}
\usepackage[T1]{fontenc}

\usepackage{amsmath,amsfonts,bm}

\def\eqref#1{equation~\ref{#1}}

\def\1{\bm{1}}

\DeclareMathAlphabet{\mathsfit}{\encodingdefault}{\sfdefault}{m}{sl}
\SetMathAlphabet{\mathsfit}{bold}{\encodingdefault}{\sfdefault}{bx}{n}

\usepackage{amsmath,amssymb}
\usepackage{booktabs,graphicx,microtype,multirow,wrapfig,needspace}
\usepackage{placeins}
\usepackage{tabularx,colortbl}
\usepackage[dvipsnames]{xcolor}
\usepackage{natbib}
\usepackage[hyphens]{url}
\usepackage{enumitem}
\hypersetup{colorlinks=true,linkcolor=black,citecolor=blue,urlcolor=blue}
\newcommand{\NA}{\mbox{\textcolor{gray}{\ensuremath{-}}}}

\newcolumntype{Y}{>{\raggedright\arraybackslash}X}

\title{Beneath the Surface: What We Overlook When We Talk About Agent Evolution}
\hypersetup{pdftitle={Beneath the Surface: What We Overlook When We Talk About Agent Evolution},pdfauthor={Xiaobo Wang, Zimu Wang, Siyu Ye, Jiaqi Li, Min Tang, Jun Bai, Yuling Shi, Ruobing Wang, Jingjing Zhang, Chaofan Wang, Kai WU, Mengnan Qi, Kai Cai, Xiaodong Gu, Qi Liu, Zilong Zheng}}
\author{\mbox{Xiaobo~Wang\textsuperscript{1,4,*}}, \mbox{Zimu~Wang\textsuperscript{2,*}}, \mbox{Siyu~Ye\textsuperscript{3,*}}, \mbox{Jiaqi~Li\textsuperscript{4,*}}, \mbox{Min~Tang\textsuperscript{1}}, \mbox{Jun~Bai\textsuperscript{4}}, \mbox{Yuling~Shi\textsuperscript{5}}, \mbox{Ruobing~Wang\textsuperscript{3}}, \mbox{Jingjing~Zhang\textsuperscript{2}}, \mbox{Chaofan~Wang\textsuperscript{5}}, \mbox{Kai~WU\textsuperscript{3}}, \mbox{Mengnan~Qi\textsuperscript{3}}, \mbox{Kai~Cai\textsuperscript{3}}, \mbox{Xiaodong~Gu\textsuperscript{5,\ensuremath{\dagger}}}, \mbox{Qi~Liu\textsuperscript{1,\ensuremath{\dagger}}}, \mbox{Zilong~Zheng\textsuperscript{4,\ensuremath{\dagger}}}}
\affiliations{
\mbox{\textsuperscript{1}University of Science and Technology of China}\quad\mbox{\textsuperscript{2}University of California, Berkeley}\\
\mbox{\textsuperscript{3}Simple Agent Lab}\quad\mbox{\textsuperscript{4}Beijing Institute for General Artificial Intelligence}\\
\mbox{\textsuperscript{5}Shanghai Jiaotong University}\\[0.4em]
\textsuperscript{*}Equal contribution.\quad\textsuperscript{\ensuremath{\dagger}}Corresponding authors.}
\date{}

\begin{document}
\maketitle
\thispagestyle{firstpagestyle}
\begin{abstract}
Agent harness evolution enables recursive self-improvement (RSI) without updating model weights. We examine whether higher benchmark scores yield reliable, efficient, and transferable improvement. We evaluate four methods across harnesses, general and coding tasks. Repeated evaluations, controlled agent expansion, component interventions, and admission replays reveal recurring patterns. All sixteen evolved agents gain previously uncovered tasks while losing previously covered ones. In one general-task case, reliability rises by 5.2 points despite losing ten previously covered tasks. Most reliability gains also increase deployment cost. Specific changes usually outperform generic reminders, but their benefits can change when combined. The highest-scoring agent does not yield the largest mean improvement in any of our four pools, and stricter admission can discard useful alternatives. Transfer selectively preserves gains: coverage usually falls on the general-task target across both harness types and paradigms, but rises on the coding target with the outer-layer-only harness. These findings motivate preservation-aware evaluation, tested decision rules, and explicit applicability conditions. As a constructive follow-up, EvoAudit turns these diagnostics into revisions of \emph{select}, \emph{analyze}, \emph{mutate}, and \emph{gate}, achieving the best performance.

\medskip\noindent{\fontfamily{phv}\selectfont\bfseries Code:} \url{https://github.com/simple-agent-lab/RSIHub}\hfill{\small\color{gray}\today}
\par\smallskip
{\noindent{\fontfamily{phv}\selectfont\bfseries Correspondence:} Siyu Ye at \href{mailto:andyyesiyu@gmail.com}{\nolinkurl{andyyesiyu@gmail.com}}, Xiaodong Gu at \href{mailto:xiaodong.gu@sjtu.edu.cn}{\nolinkurl{xiaodong.gu@sjtu.edu.cn}}, Jiaqi Li at \href{mailto:lijiaqi@bigai.ai}{\nolinkurl{lijiaqi@bigai.ai}}, Zilong Zheng at \href{mailto:zlzheng@bigai.ai}{\nolinkurl{zlzheng@bigai.ai}}, Xiaobo Wang at \href{mailto:xiaobow.0519@gmail.com}{\nolinkurl{xiaobow.0519@gmail.com}}.\par}
\end{abstract}

\section{Introduction}
\label{sec:introduction}

Recursive self-improvement (RSI) studies agents that iteratively revise their own designs based on experience \citep{zelikman2024stop,hu2025adas,zhang2026dgm}. Agent harness evolution offers a concrete form of this idea for language-model agents: it keeps model weights fixed while persistently changing the surrounding system that shapes observations and turns model outputs into actions \citep{lin2026ahe,zhang2026hyperagents}. Depending on the harness, these changes can target prompts, skills, tool interfaces, orchestration, or source code, going beyond static prompting and memory-based adaptation \citep{shinn2023reflexion,wang2024voyager}. Recent frameworks close the loop between execution, diagnosis, mutation, and evaluation, reporting gains on standard benchmarks \citep{hu2025adas,shang2025agentsquare,zhang2025aflow}. These benchmark gains suggest that harness evolution is a promising avenue for pursuing RSI.

However, a higher overall benchmark score do not mean a more useful agent. It averages over tasks and trials, so newly solved tasks can conceal lost ones, while deployment cost and performance beyond the tasks that guided evolution remain unseen \citep{gao2026evoagentbench,wang2026rethinkingharnessevolution,kapoor2025agents}. Relying on that score alone can favor brittle or expensive agent versions and mistake adaptation to the tasks used during evolution for general improvement. Meaningful progress instead requires gains that are reliable across tasks, affordable to deploy, and effective on tasks that did not guide evolution. We aim to explain which persistent harness changes produce useful gains and how evolution generates and preserves them.

We organize our empirical investigation around three research questions:

\begin{itemize}[leftmargin=*,itemsep=0pt,parsep=0pt,topsep=2pt,partopsep=0pt]
    \item \textbf{RQ1 (Phenomenology):} Does agent evolution yield reliable and efficient improvements, or does it suffer from task-coverage trade-offs and hidden costs?
    \item \textbf{RQ2 (Generalization):} Do the gains achieved during evolving survive when the agent is deployed in novel environments or tasks?
    \item \textbf{RQ3 (Mechanisms):} How do specific evolutionary operators contribute to the generation and preservation of beneficial modifications?
\end{itemize}

We study AHE \citep{lin2026ahe}, HyperAgents \citep{zhang2026hyperagents}, A-Evolve \citep{lin2026aevolve}, and GEPA \citep{agrawal2026gepa} with MiniSWE \citep{yang2024sweagent} and Codex \citep{openai_codex} on $\tau^3$-Banking \citep{shi2026tauknowledge} for general tasks and Terminal-Bench~2 \citep{merrill2026terminalbench} for coding tasks. Repeated evaluations compare evolved and initial agents in reliability, empirical coverage, regressions, and deployment cost. Split and cross-benchmark comparisons assess generalization, while controlled expansion of existing agents, change interventions, and admission replays trace how gains arise and persist.

\begin{figure}[htbp]
\centering
\includegraphics[width=0.94\textwidth]{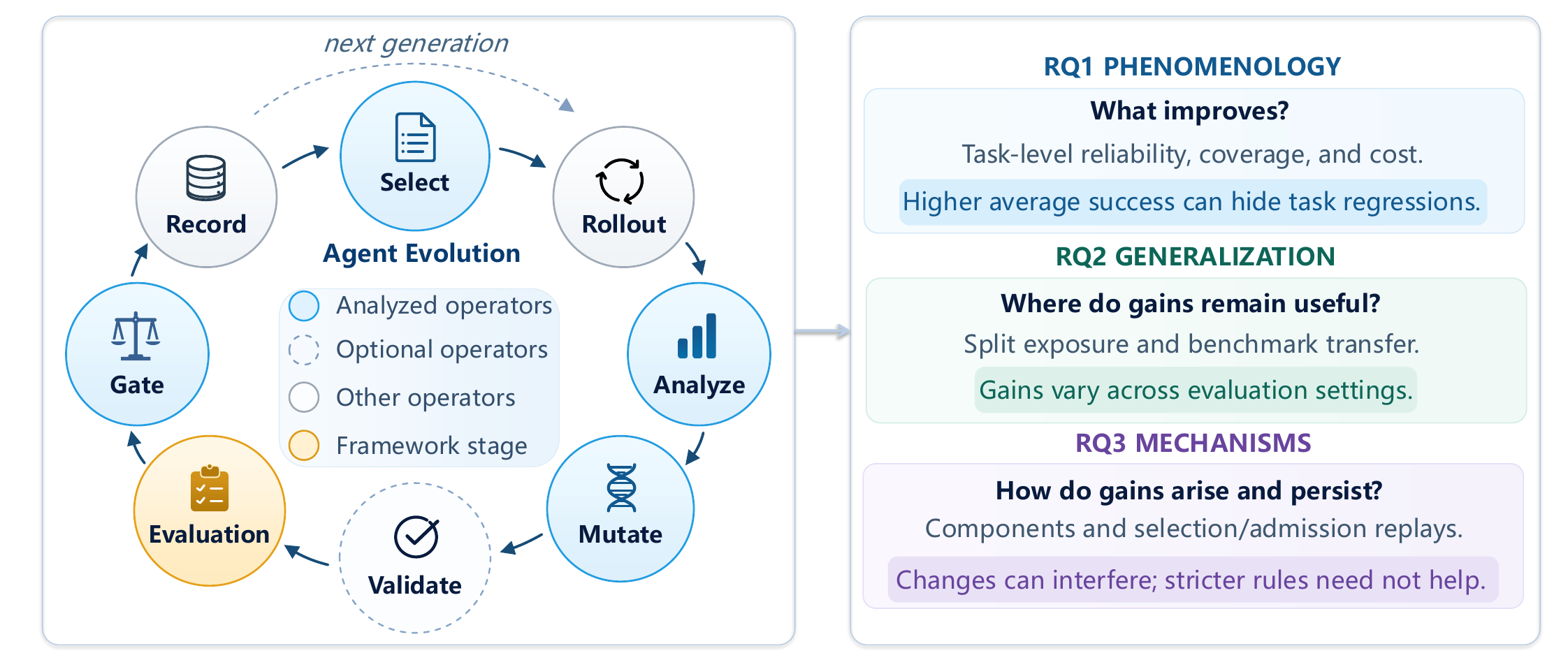}
\caption{The agent harness evolution procedure (left) and our analyses of Phenomenology, Generalization, and Mechanisms (right).}
\label{fig:analysis-framework}
\end{figure}

Harness evolution redistributes task success, with new coverage accompanied by regressions across harnesses, task domains, and evolution paradigms. The practical value of these gains depends on deployment cost and whether they persist beyond the tasks that guided evolution. Specific changes usually outperform generic reminders, yet combining them can alter their value. The strongest existing agent need not offer the greatest further improvement, and stricter admission can discard useful alternatives. Useful improvement therefore requires productive starting points, compatible changes, and search decisions that preserve gains.
 
We connect these findings to operator redesign as a constructive follow-up. Analyses of Phenomenology, Generalization, and Mechanisms jointly guide revision hypotheses for \emph{select}, \emph{analyze}, \emph{mutate}, and \emph{gate}. Each hypothesis specifies its supporting evidence, expected effects, and applicability conditions. We combine these revisions into \emph{Audit-Guided Evolution} (AGE) to test their practical value.

Our contributions can be summarized as:

\begin{itemize}[leftmargin=*,itemsep=0pt,parsep=0pt,topsep=2pt,partopsep=0pt]
    \item \textbf{Revisiting Practical Improvement.} We reveal a gap between benchmark gains and practical utility, showing that useful improvement in agent harness evolution requires preserving useful behavior, managing deployment cost, and sustaining gains beyond the tasks that guide evolution.
    \item \textbf{Understanding the Fragility of Gains.} We identify productive agent selection, change compatibility, and search retention as central challenges. Current performance can misrepresent further improvement potential, useful changes can interfere, and admission can discard stronger versions.
    \item \textbf{Improving Evolution Through Analysis.} We translate these findings into coordinated revisions of diagnosis, mutation, selection, and admission. The resulting AGE algorithm improves reliability and coverage over its reference method across both benchmarks and harnesses.
\end{itemize}

\section{Related Work}
\label{sec:related}

\textbf{Agent Optimization and Evolution.} Agent improvement spans inference-time refinement, optimization of persistent artifacts, and open-ended self-modification. Reflexion retains linguistic feedback in episodic memory across trials, whereas Self-Refine iteratively revises the current output \citep{shinn2023reflexion,madaan2023selfrefine}. ReflectEvo trains small language models on self-generated reflections to improve introspection and reasoning \citep{li2025reflectevo}. RAM learns from communications to improve an external memory system \citep{li2024ram}, and The AI Hippocampus surveys implicit, explicit, and agentic memory mechanisms \citep{jia2025aihippocampus}. DSPy optimizes language-model programs, and TextGrad propagates textual feedback through computational graphs \citep{khattab2024dspy,yuksekgonul2025textgrad}. Promptbreeder evolves both task and mutation prompts, ACE maintains evolving context playbooks, and CoEvoSkills co-evolves multi-file skill packages with a separate surrogate verifier \citep{fernando2024promptbreeder,zhang2026ace,zhang2026coevoskills}. The four approaches examined here make different choices within this space. AHE exposes editable harness components and uses distilled execution evidence to guide revisions \citep{lin2026ahe}. HyperAgents makes both the task agent and its modification procedure editable \citep{zhang2026hyperagents}. A-Evolve frames deployment-time adaptation as goal-directed optimization over persistent system state \citep{lin2026aevolve}. GEPA combines trajectory reflection for prompt updates with Pareto selection of complementary candidates \citep{agrawal2026gepa}.

\textbf{Harness and Program Evolution.} Related approaches also optimize system structure and executable code. GPTSwarm optimizes prompt-bearing graph nodes and orchestration edges, while AgentSquare searches combinations of planning, reasoning, tool-use, and memory modules \citep{zhuge2024gptswarm,shang2025agentsquare}. ADAS uses a meta agent to generate agent programs \citep{hu2025adas}. Self-modification extends to the improvement procedure itself: STOP revises optimizer scaffolding, and the Darwin G\"odel Machine evolves its own coding agent \citep{zelikman2024stop,zhang2026dgm}. AlphaEvolve instead evolves target algorithm programs using executable evaluation feedback \citep{novikov2025alphaevolve}. Recent work also explores optimizers that construct their own optimization pipelines \citep{xue2026oeo}. RSIHub composes evolution methods from reusable operators and provides shared evaluation and lineage tracking \citep{wang2026rsihub}. This distinction between improving task solutions, agent implementations, and optimization procedures motivates separating what changes from how those changes are generated and retained.

\textbf{Capability Boundaries and Mechanisms.} Empirical studies of reasoning distinguish improved small-$k$ success from expansion of large-$k$ task coverage \citep{yue2025rlboundary}. Token-level analyses identify concentrated learning signals at high-entropy decisions \citep{wang2025entropy}, while studies of on-policy distillation examine behavioral compatibility and information novelty \citep{li2026opd}. Related work also examines the feedback used for improvement. UAPO introduces uncertainty-aware utility anchors for preference optimization \citep{wang2025uapo}, and SAVE uses value-anchored on-policy feedback to update reward models \citep{wang2026flip}. We apply a related analytical perspective to persistent agent-system changes rather than model-weight updates. Repeated evaluations distinguish reliability from empirical coverage, common-process Parent expansion tests improvement potential, Component interventions examine change compatibility, and frozen-pool admission replays assess retention.

\textbf{Evaluation of Agent Systems.} Agent evaluation must account for cost, reproducibility, and generalization beyond aggregate success \citep{kapoor2025agents}. For harness evolution specifically, \citet{wang2026rethinkingharnessevolution} find inconsistent advantages over test-time scaling under comparable feedback and inference budgets, together with limited held-out generalization. Performance also depends on interactions among the model, harness, task structure, and coordination overhead \citep{kim2026scaling}. The $\tau$-bench family emphasizes reliable tool-agent-user interaction, while Terminal-Bench~2 evaluates long-horizon execution in terminal environments \citep{yao2024taubench,merrill2026terminalbench}. We study general and coding tasks through these complementary settings, distinguish task exposure during evolution, and assess external transfer without further agent revision. These comparisons connect measured improvement to its deployment cost and scope of applicability.

\section{Preliminaries}
\label{sec:framework}

\subsection{Definitions}
\label{sec:preliminaries}

An agent $A=(M,H)$ consists of a frozen model $M$ and a harness $H$. Following \citet{lin2026ahe,agrawal2026gepa}, an evolution algorithm $E$ starts from a \emph{Seed} $A_0=(M,H_0)$ and runs for $G$ generations. At generation $g$, a \emph{Parent} is an eligible existing agent selected for modification, and $\mathcal{C}_g$ contains the resulting proposed harnesses. Each $H\in\mathcal{C}_g$ defines a \emph{Candidate} agent $(M,H)$. The final \emph{Champion} is $A_\star=(M,H_\star)$:
\begingroup
\setlength{\abovedisplayskip}{5pt plus 1pt minus 2pt}
\setlength{\belowdisplayskip}{5pt plus 1pt minus 2pt}
\begin{equation}
\mathcal{C}_g=E(H_{<g},\mathcal{Z}_{<g},\mathcal{D}_{\mathrm{evo}};B_{\mathrm{evo}}),
\quad
H_\star=Q\!\left(\{H_0\}\cup\bigcup_{g=1}^{G}\mathcal{C}_g\right),
\label{eq:agent-evolution}
\end{equation}
where $H_{<g}$ is the history of harness versions before generation $g$, $\mathcal{Z}_{<g}$ is their evaluation evidence, $\mathcal{D}_{\mathrm{evo}}$ is the task set available during evolution, and $B_{\mathrm{evo}}$ is the search budget. The rule $Q$ selects the final Champion harness from the archived versions.
\endgroup

\subsection{Experiment Setup}

\subsubsection{Evolution Process and Methods}
\label{sec:analysis-procedure}
\textbf{Methods and Harnesses.} We evaluate \textbf{AHE} \citep{lin2026ahe}, \textbf{HyperAgents} \citep{zhang2026hyperagents}, \textbf{A-Evolve} \citep{lin2026aevolve}, and \textbf{GEPA} \citep{agrawal2026gepa} using \textbf{GPT-5.4}. AHE and HyperAgents are \emph{implementation-centric} recipes that emphasize agent engineering, whereas A-Evolve and GEPA are \emph{reflection-centric} recipes that distill execution feedback into reusable changes. We run each recipe with two harnesses. \textbf{MiniSWE 2.4.5} \citep{yang2024sweagent} is \emph{source-editable} and permits inner-source changes. \textbf{Codex 0.143.0} \citep{openai_codex} is \emph{outer-layer-only} and permits changes to prompts, skills, configuration, and external orchestration. These combinations let us examine different evolution designs under a common process. Appendix~\ref{app:operator-implementations} details their implementations.

In this common process, each recipe specifies how seven operators are implemented: \emph{select} chooses a Parent, \emph{rollout} collects execution evidence, \emph{analyze} diagnoses it, and \emph{mutate} proposes a Candidate. The \emph{validate} operator optionally checks the proposal, \emph{gate} decides admission, and \emph{record} archives outcomes. We focus the mechanism analysis on \emph{select}, \emph{analyze}, \emph{mutate}, and \emph{gate} because their decisions determine which Parents are used, how evidence guides changes, and which Candidates persist.

Figure~\ref{fig:analysis-framework} connects this operator process to the study's empirical comparisons. The Phenomenology analysis characterizes gains and regressions through repeated task evaluations. The Generalization analysis examines how these gains vary across original splits and external benchmarks. The Mechanisms analysis examines Parent improvement potential under a common generation process, Component interactions, and admission decisions on frozen pools.

\subsubsection{Benchmarks and Evaluation Metrics}
\label{sec:prelim-metrics}

\textbf{Benchmarks.} General tasks are represented by \textbf{$\tau^3$-Banking} \citep{shi2026tauknowledge}, with 97 tasks, and coding tasks by \textbf{Terminal-Bench 2} \citep{merrill2026terminalbench}, with 89 tasks. Both are split into Train/Validation/Test. Evolution uses Train/Validation evidence for search and admission, while Test is reserved for Seed and Champion evaluation.

\begingroup
\setlength{\abovedisplayskip}{5pt plus 1pt minus 2pt}
\setlength{\belowdisplayskip}{5pt plus 1pt minus 2pt}
\textbf{Performance.} Let $\mathcal{D}$ be an evaluation task set with $N=|\mathcal{D}|$ tasks, each executed once per pass over $k$ passes. The outcome $Y_{ni}(A)\in\{0,1\}$ indicates whether agent $A$ succeeds on task $n$ in pass $i$. With $\mathbb{I}[\cdot]$ an indicator function, \emph{Reliability} (Avg@k) and \emph{Coverage} (Pass@k) are:
\begin{equation}
\mathrm{Avg}@k = \frac{1}{Nk}\sum_{n=1}^{N}\sum_{i=1}^{k}Y_{ni}(A),
\quad
\mathrm{Pass}@k = \frac{1}{N}\sum_{n=1}^{N}\mathbb{I}\!\left[\max_{1\leq i\leq k}Y_{ni}(A)=1\right].
\label{eq:success-metrics}
\end{equation}
\emph{Task turnover} counts coverage gains and losses relative to Seed. $\mathrm{Std}@k$ is the population standard deviation of per-pass success rates.

\textbf{Cost.} Each model call $q$ costs $c(q)$. A task execution $t$ costs $c_t$, the sum of these charges over the agent's calls. Evolution cost $C_{\mathrm{evo}}$ sums costs over the set of evolution calls $\mathcal{L}_{\mathrm{evo}}$. For deployment, $\mathcal{T}_k$ contains executions of $A$ on $\mathcal{D}$ with recorded costs in the first $k$ passes. Their mean cost is $\bar c_{\mathrm{dep}}^{(k)}$. Multiplying it by $k$ gives the estimated per-task deployment cost $C_{\mathrm{dep}}@k$:
\begin{equation}
C_{\mathrm{evo}}=\sum_{q\in\mathcal{L}_{\mathrm{evo}}}c(q),
\quad
\bar c_{\mathrm{dep}}^{(k)}(A;\mathcal{D})=\frac{1}{|\mathcal{T}_k|}\sum_{t\in\mathcal{T}_k}c_t,
\quad
C_{\mathrm{dep}}@k(A;\mathcal{D})=k\,\bar c_{\mathrm{dep}}^{(k)}(A;\mathcal{D}).
\label{eq:cost}
\end{equation}
\endgroup

\section{Phenomenology: Revisiting What Has Been Evolved}
\label{sec:main-results}

In this section, we compare each frozen Champion with its matching Seed in reliability, empirical coverage, task-level transitions, and deployment cost. Table~\ref{tab:overall} reports eight-pass aggregate results, Figure~\ref{fig:task-transitions} shows paired task-level success counts, and Figure~\ref{fig:deployment-cost-curves} relates $\mathrm{Pass}@k$ to $C_{\mathrm{dep}}@k$ for $k=1,2,4,8$ using the same canonical trials. Appendix~\ref{app:rq1-details} provides detailed comparisons and stability checks.

\subsection{Overall Reliability and Empirical Task Coverage}

\begin{table}[htbp]
\caption{Reliability and empirical coverage across eight passes. Generation 0 is the Seed; later entries give the selected Champion's Generation. Scores are percentages and deltas are percentage points.}
\label{tab:overall}
\centering
\resizebox{0.9\textwidth}{!}{%
\begin{tabular}{lllccccc}
\toprule
Benchmark & Harness & Variant & Generation & $\mathrm{Avg}@8\pm\mathrm{Std}$ & $\Delta\mathrm{Avg}$ & $\mathrm{Pass}@8$ & $\Delta\mathrm{Pass}$ \\
\midrule
\multirow{10}{*}{$\tau^3$-Banking} & \multirow{5}{*}{MiniSWE}
& Seed         & 0  & $27.4\!\pm\!3.1$ & \NA & $54.6$ & \NA \\
& & AHE          & 2  & $32.9\!\pm\!3.9$ & \textcolor{green!50!black}{$+5.4$} & $66.0$ & \textcolor{green!50!black}{$+11.3$} \\
& & HyperAgents  & 9  & $32.6\!\pm\!4.1$ & \textcolor{green!50!black}{$+5.2$} & $57.7$ & \textcolor{green!50!black}{$+3.1$} \\
& & A-Evolve     & 1  & $26.3\!\pm\!2.5$ & \textcolor{red!75!black}{$-1.2$} & $55.7$ & \textcolor{green!50!black}{$+1.0$} \\
& & GEPA         & 11 & $28.2\!\pm\!2.3$ & \textcolor{green!50!black}{$+0.8$} & $58.8$ & \textcolor{green!50!black}{$+4.1$} \\
\cmidrule(l){2-8}
& \multirow{5}{*}{Codex}
& Seed         & 0  & $30.3\!\pm\!4.3$ & \NA & $56.7$ & \NA \\
& & AHE          & 15 & $36.2\!\pm\!3.4$ & \textcolor{green!50!black}{$+5.9$} & $62.9$ & \textcolor{green!50!black}{$+6.2$} \\
& & HyperAgents  & 18 & $36.9\!\pm\!2.5$ & \textcolor{green!50!black}{$+6.6$} & $66.0$ & \textcolor{green!50!black}{$+9.3$} \\
& & A-Evolve     & 1  & $28.1\!\pm\!2.9$ & \textcolor{red!75!black}{$-2.2$} & $57.7$ & \textcolor{green!50!black}{$+1.0$} \\
& & GEPA         & 5  & $29.5\!\pm\!3.2$ & \textcolor{red!75!black}{$-0.8$} & $53.6$ & \textcolor{red!75!black}{$-3.1$} \\
\midrule
\multirow{10}{*}{Terminal-Bench~2} & \multirow{5}{*}{MiniSWE} & Seed & 0 & $65.9\!\pm\!4.8$ & \NA & $86.5$ & \NA \\
 &  & AHE & 18 & $70.2\!\pm\!2.3$ & \textcolor{green!50!black}{$+4.4$} & $87.6$ & \textcolor{green!50!black}{$+1.1$} \\
 &  & HyperAgents & 17 & $67.7\!\pm\!3.2$ & \textcolor{green!50!black}{$+1.8$} & $85.4$ & \textcolor{red!75!black}{$-1.1$} \\
 &  & A-Evolve & 4 & $68.0\!\pm\!2.9$ & \textcolor{green!50!black}{$+2.1$} & $86.5$ & $0.0$ \\
 &  & GEPA & 20 & $68.5\!\pm\!3.5$ & \textcolor{green!50!black}{$+2.7$} & $88.8$ & \textcolor{green!50!black}{$+2.2$} \\
\cmidrule(l){2-8}
 & \multirow{5}{*}{Codex} & Seed & 0 & $69.9\!\pm\!2.3$ & \NA & $89.9$ & \NA \\
 &  & AHE & 20 & $72.1\!\pm\!3.4$ & \textcolor{green!50!black}{$+2.1$} & $91.0$ & \textcolor{green!50!black}{$+1.1$} \\
 &  & HyperAgents & 18 & $72.6\!\pm\!3.9$ & \textcolor{green!50!black}{$+2.7$} & $86.5$ & \textcolor{red!75!black}{$-3.4$} \\
 &  & A-Evolve & 2 & $68.0\!\pm\!1.7$ & \textcolor{red!75!black}{$-2.0$} & $87.6$ & \textcolor{red!75!black}{$-2.2$} \\
 &  & GEPA & 15 & $68.1\!\pm\!4.7$ & \textcolor{red!75!black}{$-1.8$} & $86.5$ & \textcolor{red!75!black}{$-3.4$} \\
\bottomrule
\end{tabular}}
\par
\end{table}

\textbf{Aggregate Improvement Is Not a Single Outcome.} Table~\ref{tab:overall} shows reliability and coverage gains in eleven and ten of sixteen lineages, respectively. Four move the metrics in opposite directions across both harnesses, domains, and paradigms; another improves only reliability. Consistency and empirical reach are therefore distinct outcomes that aggregate improvement claims should separate.

\subsection{Coverage Is a Flow}

\begin{figure}[!htbp]
\centering
\includegraphics[width=0.85\textwidth]{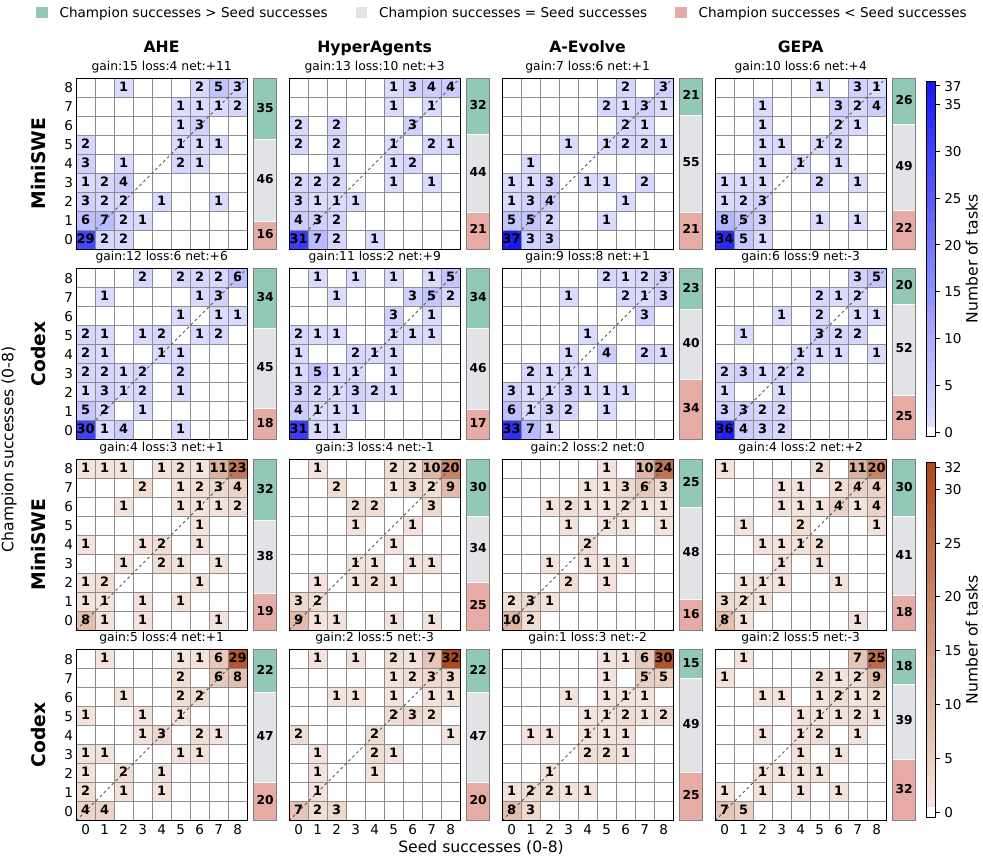}
\caption{Task-level Seed-to-Champion transitions on $\tau^3$-Banking (blue) and Terminal-Bench~2 (orange). Each cell reports the number of tasks with the corresponding Seed and Champion success counts. Right-hand bars group tasks by whether the Champion succeeds more than, the same as, or less than the Seed. Non-origin cells on the left or bottom edges indicate gained or lost coverage.}
\label{fig:task-transitions}
\end{figure}

\textbf{Evolution Redistributes Success.} All sixteen Champions in Figure~\ref{fig:task-transitions} gain and lose covered tasks, and increase repeated success on some tasks while decreasing it on others. This pattern spans both harnesses, domains, and paradigms: even when both aggregate metrics improve, coverage does not expand monotonically. Gains and losses should therefore be reported separately. Appendix~\ref{app:rq1-transitions} gives further analysis.

\textbf{Behavioral Changes Behind Task-Level Gains and Losses.} We inspect contrasting traces for tasks with large success-count changes. Across the sampled harnesses, domains, and paradigms, recurring differences concern state preservation, precise grounding, and task completion. Successful traces preserve recovery files and respect action dependencies, while failures include incorrect arguments, premature termination, and solutions that work locally but fail in the required environment. These contrasts motivate preservation and end-to-end checks rather than equating every outcome flip with acquired or lost task knowledge. Table~\ref{tab:behavior-audit-summary} summarizes the categories; Appendix~\ref{app:rq1-behavior} gives case evidence.

\begin{table}[htbp]
\centering
\caption{Primary behavior categories in the 22 selected task--lineage comparisons. Increase and decrease refer to eight-pass success counts.}
\label{tab:behavior-audit-summary}
\small
\begin{tabularx}{\textwidth}{@{}m{0.25\textwidth}Ycc@{}}
\toprule
Behavior category & Observable distinction & Increase & Decrease \\
\midrule
State and dependency management & Preserve inputs, respect action dependencies, and retain a self-contained executable result. & 7 & 1 \\
Exact grounding & Match product conditions, dates, or output characters to the evidence and task requirements. & 1 & 4 \\
Completion and handoff & Finish all requested work and deliver the required final state or actual handoff. & 2 & 3 \\
Tool discovery and recovery & Use an available interface and recover from tool errors using the current state. & 2 & 1 \\
Implementation debugging & Resolve implementation errors and deliver a working program within the execution window. & 1 & 0 \\
\bottomrule
\end{tabularx}
\end{table}

\subsection{Reliability and Deployment Efficiency}

\begin{figure}[htbp]
\centering
\includegraphics[width=\textwidth]{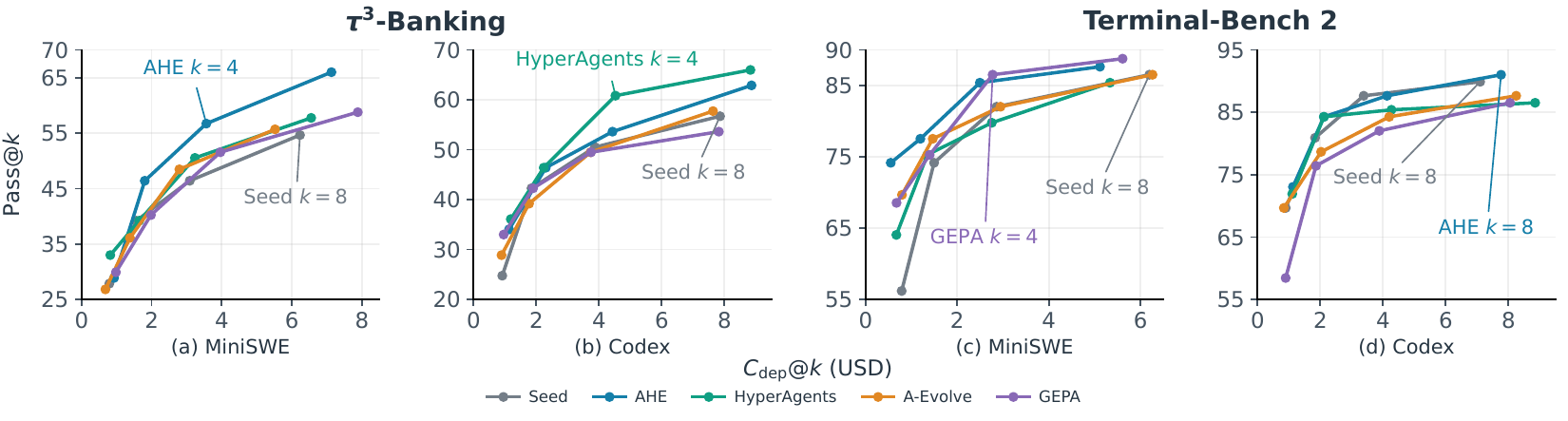}
\caption{Coverage--cost curves for MiniSWE and Codex on both benchmarks. Points pair $\mathrm{Pass}@k$ with $C_{\mathrm{dep}}@k$ from Equation~\ref{eq:cost} for $k=1,2,4,8$.}
\label{fig:deployment-cost-curves}
\end{figure}

\textbf{Per-Trial Cost and Deployment Efficiency Can Diverge.} Eight of eleven reliability-improving Champions cost more per trial than Seed; the three lower-cost gains use the source-editable harness on coding tasks. Yet Figure~\ref{fig:deployment-cost-curves} shows that higher per-trial cost can coexist with fewer attempts and lower cost to reach comparable coverage. MiniSWE with AHE on $\tau^3$-Banking reaches 56.7\% coverage in four passes, exceeding Seed's 54.6\% in eight at lower deployment cost. Practical efficiency therefore depends on both cost per attempt and the attempts needed to attain a given coverage level.

\section{Generalization: Regularizing The Benefits Of Evolution Within A Controllable Boundary}
\label{sec:transfer}

Table~\ref{tab:ood-profile} compares frozen Champions with matching Seeds across evolution splits and external transfer. Train, Validation, and Test report contributions to the full source-benchmark gain, summing to the overall change before rounding. Transfer reports gains measured on each external benchmark relative to its matching Seed. Appendix~\ref{app:rq2-details} gives protocols and interpretation.

\begin{table}[htbp]
\caption{Seed-relative gains across evolution splits and external transfer in percentage points. Darker cells indicate larger gains on a shared linear scale.}
\label{tab:ood-profile}
\label{tab:split-gains}
\centering
\footnotesize
\setlength{\tabcolsep}{3pt}
\resizebox{0.9\textwidth}{!}{%
\begin{tabular}{@{}lll*{4}{c}@{\hspace{8pt}}*{4}{c}@{}}
\toprule
\multirow{3}{*}{Benchmark} & \multirow{3}{*}{Harness} & \multirow{3}{*}{Method} & \multicolumn{4}{c}{$\Delta\mathrm{Pass}@8$} & \multicolumn{4}{c}{$\Delta\mathrm{Avg}@8$} \\
\cmidrule(lr){4-7}\cmidrule(lr){8-11}
& & & \multicolumn{4}{c}{$\xrightarrow{\text{From evolution data to OOD transfer}}$} & \multicolumn{4}{c}{$\xrightarrow{\text{From evolution data to OOD transfer}}$} \\
& & & Train & Validation & Test & Transfer & Train & Validation & Test & Transfer \\
\midrule
\multirow{8}{*}{$\tau^3$-Banking} & \multirow{4}{*}{MiniSWE} & AHE & \cellcolor[HTML]{749EC9}\textcolor{black}{$+3.1$} & \cellcolor[HTML]{5F8FC0}\textcolor{black}{$+4.1$} & \cellcolor[HTML]{5F8FC0}\textcolor{black}{$+4.1$} & \cellcolor[HTML]{C4D7EA}\textcolor{black}{$-0.7$} & \cellcolor[HTML]{6E99C6}\textcolor{black}{$+3.4$} & \cellcolor[HTML]{A2BFDC}\textcolor{black}{$+0.9$} & \cellcolor[HTML]{9CBAD9}\textcolor{black}{$+1.2$} & \cellcolor[HTML]{7CA4CC}\textcolor{black}{$+2.7$} \\
 &  & HyperAgents & \cellcolor[HTML]{89ADD1}\textcolor{black}{$+2.1$} & \cellcolor[HTML]{A0BDDB}\textcolor{black}{$+1.0$} & \cellcolor[HTML]{B5CCE4}\textcolor{black}{$0.0$} & \cellcolor[HTML]{BED2E7}\textcolor{black}{$-0.4$} & \cellcolor[HTML]{6996C4}\textcolor{black}{$+3.6$} & \cellcolor[HTML]{91B3D5}\textcolor{black}{$+1.7$} & \cellcolor[HTML]{B8CEE5}\textcolor{black}{$-0.1$} & \cellcolor[HTML]{A5C0DD}\textcolor{black}{$+0.8$} \\
 &  & A-Evolve & \cellcolor[HTML]{B5CCE4}\textcolor{black}{$0.0$} & \cellcolor[HTML]{A0BDDB}\textcolor{black}{$+1.0$} & \cellcolor[HTML]{B5CCE4}\textcolor{black}{$0.0$} & \cellcolor[HTML]{BCD1E6}\textcolor{black}{$-0.3$} & \cellcolor[HTML]{D9E6F3}\textcolor{black}{$-1.7$} & \cellcolor[HTML]{ADC6E0}\textcolor{black}{$+0.4$} & \cellcolor[HTML]{B3CBE3}\textcolor{black}{$+0.1$} & \cellcolor[HTML]{98B7D8}\textcolor{black}{$+1.4$} \\
 &  & GEPA & \cellcolor[HTML]{A0BDDB}\textcolor{black}{$+1.0$} & \cellcolor[HTML]{89ADD1}\textcolor{black}{$+2.1$} & \cellcolor[HTML]{A0BDDB}\textcolor{black}{$+1.0$} & \cellcolor[HTML]{96B6D7}\textcolor{black}{$+1.5$} & \cellcolor[HTML]{ADC6E0}\textcolor{black}{$+0.4$} & \cellcolor[HTML]{ABC5DF}\textcolor{black}{$+0.5$} & \cellcolor[HTML]{B8CEE5}\textcolor{black}{$-0.1$} & \cellcolor[HTML]{7EA5CD}\textcolor{black}{$+2.6$} \\
\cmidrule(l){2-11}
 & \multirow{4}{*}{Codex} & AHE & \cellcolor[HTML]{326FAE}\textcolor{white}{$+6.2$} & \cellcolor[HTML]{B5CCE4}\textcolor{black}{$0.0$} & \cellcolor[HTML]{B5CCE4}\textcolor{black}{$0.0$} & \cellcolor[HTML]{C4D7EA}\textcolor{black}{$-0.7$} & \cellcolor[HTML]{729CC8}\textcolor{black}{$+3.2$} & \cellcolor[HTML]{9CBAD9}\textcolor{black}{$+1.2$} & \cellcolor[HTML]{96B6D7}\textcolor{black}{$+1.5$} & \cellcolor[HTML]{D9E6F3}\textcolor{black}{$-1.7$} \\
 &  & HyperAgents & \cellcolor[HTML]{08519C}\textcolor{white}{$+8.2$} & \cellcolor[HTML]{A0BDDB}\textcolor{black}{$+1.0$} & \cellcolor[HTML]{B5CCE4}\textcolor{black}{$0.0$} & \cellcolor[HTML]{C4D7EA}\textcolor{black}{$-0.7$} & \cellcolor[HTML]{5F8FC0}\textcolor{black}{$+4.1$} & \cellcolor[HTML]{8DB0D3}\textcolor{black}{$+1.9$} & \cellcolor[HTML]{ABC5DF}\textcolor{black}{$+0.5$} & \cellcolor[HTML]{AFC8E1}\textcolor{black}{$+0.3$} \\
 &  & A-Evolve & \cellcolor[HTML]{A0BDDB}\textcolor{black}{$+1.0$} & \cellcolor[HTML]{89ADD1}\textcolor{black}{$+2.1$} & \cellcolor[HTML]{E2ECF6}\textcolor{black}{$-2.1$} & \cellcolor[HTML]{CBDBED}\textcolor{black}{$-1.0$} & \cellcolor[HTML]{D3E1F0}\textcolor{black}{$-1.4$} & \cellcolor[HTML]{BCD1E6}\textcolor{black}{$-0.3$} & \cellcolor[HTML]{C0D4E8}\textcolor{black}{$-0.5$} & \cellcolor[HTML]{E0EAF5}\textcolor{black}{$-2.0$} \\
 &  & GEPA & \cellcolor[HTML]{CBDBED}\textcolor{black}{$-1.0$} & \cellcolor[HTML]{A0BDDB}\textcolor{black}{$+1.0$} & \cellcolor[HTML]{F7FBFF}\textcolor{black}{$-3.1$} & \cellcolor[HTML]{CBDBED}\textcolor{black}{$-1.0$} & \cellcolor[HTML]{C8DAEC}\textcolor{black}{$-0.9$} & \cellcolor[HTML]{AFC8E1}\textcolor{black}{$+0.3$} & \cellcolor[HTML]{B8CEE5}\textcolor{black}{$-0.1$} & \cellcolor[HTML]{D3E1F0}\textcolor{black}{$-1.4$} \\
\midrule
\multirow{8}{*}{\shortstack{Terminal-\\Bench~2}} & \multirow{4}{*}{MiniSWE} & AHE & \cellcolor[HTML]{9EBCDA}\textcolor{black}{$+1.1$} & \cellcolor[HTML]{CDDDED}\textcolor{black}{$-1.1$} & \cellcolor[HTML]{9EBCDA}\textcolor{black}{$+1.1$} & \cellcolor[HTML]{8FB1D4}\textcolor{black}{$+1.8$} & \cellcolor[HTML]{78A1CA}\textcolor{black}{$+2.9$} & \cellcolor[HTML]{BED2E7}\textcolor{black}{$-0.4$} & \cellcolor[HTML]{8FB1D4}\textcolor{black}{$+1.8$} & \cellcolor[HTML]{E6EFF8}\textcolor{black}{$-2.3$} \\
 &  & HyperAgents & \cellcolor[HTML]{E4EDF7}\textcolor{black}{$-2.2$} & \cellcolor[HTML]{B5CCE4}\textcolor{black}{$0.0$} & \cellcolor[HTML]{9EBCDA}\textcolor{black}{$+1.1$} & \cellcolor[HTML]{6B98C5}\textcolor{black}{$+3.5$} & \cellcolor[HTML]{AFC8E1}\textcolor{black}{$+0.3$} & \cellcolor[HTML]{A5C0DD}\textcolor{black}{$+0.8$} & \cellcolor[HTML]{A7C2DE}\textcolor{black}{$+0.7$} & \cellcolor[HTML]{E2ECF6}\textcolor{black}{$-2.1$} \\
 &  & A-Evolve & \cellcolor[HTML]{B5CCE4}\textcolor{black}{$0.0$} & \cellcolor[HTML]{CDDDED}\textcolor{black}{$-1.1$} & \cellcolor[HTML]{9EBCDA}\textcolor{black}{$+1.1$} & \cellcolor[HTML]{588ABD}\textcolor{black}{$+4.4$} & \cellcolor[HTML]{96B6D7}\textcolor{black}{$+1.5$} & \cellcolor[HTML]{BED2E7}\textcolor{black}{$-0.4$} & \cellcolor[HTML]{A0BDDB}\textcolor{black}{$+1.0$} & \cellcolor[HTML]{BACFE6}\textcolor{black}{$-0.2$} \\
 &  & GEPA & \cellcolor[HTML]{9EBCDA}\textcolor{black}{$+1.1$} & \cellcolor[HTML]{CDDDED}\textcolor{black}{$-1.1$} & \cellcolor[HTML]{87ABD1}\textcolor{black}{$+2.2$} & \cellcolor[HTML]{EFF5FB}\textcolor{black}{$-2.7$} & \cellcolor[HTML]{9EBCDA}\textcolor{black}{$+1.1$} & \cellcolor[HTML]{B3CBE3}\textcolor{black}{$+0.1$} & \cellcolor[HTML]{98B7D8}\textcolor{black}{$+1.4$} & \cellcolor[HTML]{D9E6F3}\textcolor{black}{$-1.7$} \\
\cmidrule(l){2-11}
 & \multirow{4}{*}{Codex} & AHE & \cellcolor[HTML]{87ABD1}\textcolor{black}{$+2.2$} & \cellcolor[HTML]{E4EDF7}\textcolor{black}{$-2.2$} & \cellcolor[HTML]{9EBCDA}\textcolor{black}{$+1.1$} & \cellcolor[HTML]{0C549E}\textcolor{white}{$+8.0$} & \cellcolor[HTML]{78A1CA}\textcolor{black}{$+2.9$} & \cellcolor[HTML]{D3E1F0}\textcolor{black}{$-1.4$} & \cellcolor[HTML]{A9C3DF}\textcolor{black}{$+0.6$} & \cellcolor[HTML]{B8CEE5}\textcolor{black}{$-0.1$} \\
 &  & HyperAgents & \cellcolor[HTML]{B5CCE4}\textcolor{black}{$0.0$} & \cellcolor[HTML]{E4EDF7}\textcolor{black}{$-2.2$} & \cellcolor[HTML]{CDDDED}\textcolor{black}{$-1.1$} & \cellcolor[HTML]{326FAE}\textcolor{white}{$+6.2$} & \cellcolor[HTML]{6795C3}\textcolor{black}{$+3.7$} & \cellcolor[HTML]{D9E6F3}\textcolor{black}{$-1.7$} & \cellcolor[HTML]{A7C2DE}\textcolor{black}{$+0.7$} & \cellcolor[HTML]{749EC9}\textcolor{black}{$+3.1$} \\
 &  & A-Evolve & \cellcolor[HTML]{CDDDED}\textcolor{black}{$-1.1$} & \cellcolor[HTML]{CDDDED}\textcolor{black}{$-1.1$} & \cellcolor[HTML]{B5CCE4}\textcolor{black}{$0.0$} & \cellcolor[HTML]{8FB1D4}\textcolor{black}{$+1.8$} & \cellcolor[HTML]{B5CCE4}\textcolor{black}{$0.0$} & \cellcolor[HTML]{CBDBED}\textcolor{black}{$-1.0$} & \cellcolor[HTML]{CBDBED}\textcolor{black}{$-1.0$} & \cellcolor[HTML]{E2ECF6}\textcolor{black}{$-2.1$} \\
 &  & GEPA & \cellcolor[HTML]{CDDDED}\textcolor{black}{$-1.1$} & \cellcolor[HTML]{E4EDF7}\textcolor{black}{$-2.2$} & \cellcolor[HTML]{B5CCE4}\textcolor{black}{$0.0$} & \cellcolor[HTML]{6B98C5}\textcolor{black}{$+3.5$} & \cellcolor[HTML]{A5C0DD}\textcolor{black}{$+0.8$} & \cellcolor[HTML]{CDDDED}\textcolor{black}{$-1.1$} & \cellcolor[HTML]{D5E3F1}\textcolor{black}{$-1.5$} & \cellcolor[HTML]{98B7D8}\textcolor{black}{$+1.4$} \\
\bottomrule
\end{tabular}}
\par
\end{table}

\subsection{Generalization Within the Same Benchmarks}

\textbf{Gains Often Survive Beyond Direct Feedback.} Test reliability improves in ten of sixteen lineages, including seven of eight implementation-centric cases across both harness types and task domains. Reflection-centric gains are more harness-dependent: most source-editable cases improve, whereas all outer-layer-only cases decline. Improvement therefore frequently reaches feedback-excluded tasks, with systematic differences across harness--paradigm combinations. 

\subsection{Generalization Cross Benchmarks}

We test general-task transfer from $\tau^3$-Banking to the full $\tau^2$-bench \citep{barres2025tau2}, and coding-task transfer from Terminal-Bench~2 to DeepSWE~v1.1 \citep{huang2026deepswe}. Table~\ref{tab:ood-profile} gives Seed-relative gains and Appendix Table~\ref{tab:transfer-details} gives absolute scores.

\textbf{Transfer Selectively Preserves Gains.} Coverage falls for seven of eight Champions on the general-task target but rises for seven of eight on the coding target, spanning both harness types and paradigms. Reliability can move in the opposite direction. All source-editable Champions improve general-task transfer reliability but lose coding-task transfer reliability, despite coverage gains in three coding cases. Transfer therefore changes the balance between reliability and coverage, so the persistence of improvement must be assessed separately for each outcome and target.

\section{Mechanisms: Tracing How to Discover And Retain The Optimization Gains }
\label{sec:mechanism}

\subsection{Select: Choosing Which Agent to Improve}
\label{sec:candidate-decisions}

We test whether \emph{select} identifies Parents with the greatest improvement potential. Each Champion produces three Candidates through the same A-Evolve analysis and mutation process.

\Needspace{9\baselineskip}
\begin{wraptable}[9]{r}{0.43\textwidth}
\vspace{\dimexpr-\intextsep+4.8pt\relax}
\setlength{\abovecaptionskip}{0pt}
\setlength{\belowcaptionskip}{6pt}
\caption{Best Parent selection rates (\%). Higher values are bold.}
\label{tab:select-parent-quality}
\centering
\footnotesize
\setlength{\tabcolsep}{2pt}
\begin{tabularx}{\linewidth}{@{}>{\raggedright\arraybackslash}m{1.62cm}l>{\centering\arraybackslash}m{1.3cm}>{\centering\arraybackslash}X@{}}
\toprule
Benchmark & Harness & Original & Random \\
\midrule
\multirow{2}{*}{$\tau^3$-Banking} & MiniSWE & $17.5$ & \textbf{25.0} \\
\cmidrule(lr){2-4}
& Codex & $8.8$ & \textbf{25.0} \\
\midrule
\multirow{2}{=}{\raggedright Terminal-Bench~2} & MiniSWE & \textbf{59.5} & $25.0$ \\
\cmidrule(lr){2-4}
& Codex & \textbf{55.7} & $25.0$ \\
\bottomrule
\end{tabularx}
\par
\end{wraptable}

Each benchmark and harness supplies one four-Parent pool, with one Train pass shared by each Parent's three branches. The best Parent maximizes mean Candidate Test $\mathrm{Avg}@8$ minus its own. Table~\ref{tab:select-parent-quality} reports its selection probability. Original averages the native HyperAgents, A-Evolve, and GEPA rules using Train or Validation evidence; Random samples uniformly. Appendix~\ref{app:rq3-select-details} gives the protocol.

\textbf{Current Scores Do Not Consistently Identify Improvement Potential.} Original falls below the 25\% random baseline on general tasks but exceeds 55\% on coding tasks, across both harnesses. This reversal shows that selection signals align with further improvement differently across domains. Evaluating \emph{select} therefore requires measuring subsequent gains alongside current performance.

\subsection{Analyze: Identifying What the Evidence Supports}
\label{sec:component-mechanisms}

Tables~\ref{tab:mechanism-factorization} and~\ref{tab:generic-reminder-controls} inform \emph{analyze} by comparing eight Champions' \emph{Components}, functionally related changes with their dependencies, against \emph{generic reminders} such as checking tools and verifying results. Appendix~\ref{app:component-definitions} defines each Component, and Appendix~\ref{app:rq3-analyze-details} gives the protocol.

\begin{table}[htbp]
\caption{Isolated Components versus complete Champions. Appendix~\ref{app:component-definitions} defines each row's A and B. Scores are percentages. Best values are bold.}
\label{tab:mechanism-factorization}
\centering
\footnotesize
\begingroup
\setlength{\tabcolsep}{2pt}
\renewcommand{\arraystretch}{1.15}
\begin{tabularx}{\textwidth}{@{}>{\raggedright\arraybackslash}m{1.62cm}ll*{3}{>{\centering\arraybackslash}X}>{\centering\arraybackslash}m{1.5cm}@{\hspace{8pt}}*{3}{>{\centering\arraybackslash}X}>{\centering\arraybackslash}m{1.5cm}@{}}
\toprule
\multirow{3}{*}{Benchmark} & \multirow{3}{*}{Harness} & \multirow{3}{*}{Method} & \multicolumn{4}{c}{$\mathrm{Avg}@8$ (\%)} & \multicolumn{4}{c}{$\mathrm{Pass}@8$ (\%)} \\
\cmidrule(lr){4-7}\cmidrule(lr){8-11}
& & & Seed & \shortstack{Seed\\+ A} & \shortstack{Seed\\+ B} & Champion & Seed & \shortstack{Seed\\+ A} & \shortstack{Seed\\+ B} & Champion \\
\midrule
\multirow{4}{*}{$\tau^3$-Banking} & \multirow{2}{*}{MiniSWE} & AHE & \multirow{2}{*}{$27.4$} & $28.2$ & $32.1$ & \textbf{32.9} & \multirow{2}{*}{$54.6$} & $53.6$ & $64.9$ & \textbf{66.0} \\
 & & HyperAgents & & \textbf{34.4} & $32.0$ & $32.6$ & & \textbf{59.8} & $55.7$ & $57.7$ \\
\cmidrule(lr){2-11}
 & \multirow{2}{*}{Codex} & AHE & \multirow{2}{*}{$30.3$} & $29.6$ & $34.8$ & \textbf{36.2} & \multirow{2}{*}{$56.7$} & $56.7$ & \textbf{63.9} & $62.9$ \\
 & & HyperAgents & & \textbf{37.2} & $36.5$ & $36.9$ & & $67.0$ & \textbf{72.2} & $66.0$ \\
\midrule
\multirow{4}{1.62cm}{\raggedright Terminal-Bench~2} & \multirow{2}{*}{MiniSWE} & AHE & $65.9$ & $67.7$ & $69.1$ & \textbf{70.2} & $86.5$ & \textbf{88.8} & $87.6$ & $87.6$ \\
 & & GEPA & $65.9$ & $67.3$ & $64.0$ & \textbf{68.5} & $86.5$ & $87.6$ & $86.5$ & \textbf{88.8} \\
\cmidrule(lr){2-11}
 & \multirow{2}{*}{Codex} & AHE & $69.9$ & $70.2$ & $69.9$ & \textbf{72.1} & $89.9$ & $85.4$ & $86.5$ & \textbf{91.0} \\
 & & HyperAgents & $69.9$ & $69.1$ & $68.0$ & \textbf{72.6} & \textbf{89.9} & \textbf{89.9} & $86.5$ & $86.5$ \\
\bottomrule
\end{tabularx}
\endgroup
\par
\end{table}
\begin{table}[htbp]
\newsavebox{\reminderCaptionBox}
\newsavebox{\gateCaptionBox}
\newlength{\diagnosticCaptionHeight}
\sbox{\reminderCaptionBox}{\begin{minipage}[t]{0.55\textwidth}
\caption{Seed versus Seed with a generic reminder. Scores are percentages.}
\label{tab:generic-reminder-controls}
\end{minipage}}
\sbox{\gateCaptionBox}{\begin{minipage}[t]{0.43\textwidth}
\caption{Size-normalized best-Candidate retention (\%), averaged over four methods.}
\label{tab:gate-constraint}
\end{minipage}}
\setlength{\diagnosticCaptionHeight}{\dimexpr\ht\reminderCaptionBox+\dp\reminderCaptionBox\relax}
\ifdim\dimexpr\ht\gateCaptionBox+\dp\gateCaptionBox\relax>\diagnosticCaptionHeight
  \setlength{\diagnosticCaptionHeight}{\dimexpr\ht\gateCaptionBox+\dp\gateCaptionBox\relax}
\fi
\begin{minipage}[t]{0.55\textwidth}
\parbox[t][\diagnosticCaptionHeight][t]{\linewidth}{\usebox{\reminderCaptionBox}}\par
\centering
\footnotesize
\setlength{\tabcolsep}{2pt}
\begin{tabularx}{\linewidth}{@{}>{\raggedright\arraybackslash}m{1.62cm}l>{\centering\arraybackslash}X>{\centering\arraybackslash}m{1.3cm}@{\hspace{6pt}}>{\centering\arraybackslash}X>{\centering\arraybackslash}m{1.3cm}@{}}
\toprule
\multirow{3}{*}{Benchmark} & \multirow{3}{*}{Harness} & \multicolumn{2}{c}{$\mathrm{Avg}@8$ (\%)} & \multicolumn{2}{c}{$\mathrm{Pass}@8$ (\%)} \\
\cmidrule(lr){3-4}\cmidrule(lr){5-6}
& & \multirow{2}{*}{Seed} & Generic & \multirow{2}{*}{Seed} & Generic \\
& & & reminder & & reminder \\
\midrule
\multirow{2}{*}{$\tau^3$-Banking} & MiniSWE & $27.4$ & \textbf{27.8} & $54.6$ & \textbf{55.7} \\
\cmidrule(lr){2-6}
 & Codex & \textbf{30.3} & $28.6$ & \textbf{56.7} & \textbf{56.7} \\
\midrule
\multirow{2}{=}{\raggedright Terminal-Bench~2} & MiniSWE & \textbf{65.9} & \textbf{65.9} & \textbf{86.5} & $84.3$ \\
\cmidrule(lr){2-6}
 & Codex & \textbf{69.9} & $65.7$ & \textbf{89.9} & $88.8$ \\
\bottomrule
\end{tabularx}
\par
\end{minipage}\hfill
\begin{minipage}[t]{0.43\textwidth}
\parbox[t][\diagnosticCaptionHeight][t]{\linewidth}{\usebox{\gateCaptionBox}}\par
\centering
\footnotesize
\setlength{\tabcolsep}{2pt}
\begin{tabularx}{\linewidth}{@{}>{\raggedright\arraybackslash}m{1.62cm}l>{\centering\arraybackslash}m{1.3cm}>{\centering\arraybackslash}X@{}}
\toprule
\multirow{3}{*}{Benchmark} & \multirow{3}{*}{Harness} & \multicolumn{2}{c}{Retention (\%)} \\
\cmidrule(lr){3-4}
& & \multirow{2}{*}{Original} & \multirow{2}{*}{Constrained} \\
& & & \\
\midrule
\multirow{2}{*}{$\tau^3$-Banking} & MiniSWE & \textbf{31.3} & $25.0$ \\
\cmidrule(lr){2-4}
 & Codex & $18.8$ & \textbf{25.0} \\
\midrule
\multirow{2}{=}{\raggedright Terminal-Bench~2} & MiniSWE & \textbf{25.0} & $12.5$ \\
\cmidrule(lr){2-4}
 & Codex & $43.8$ & \textbf{50.0} \\
\bottomrule
\end{tabularx}
\par
\end{minipage}
\end{table}

\textbf{Specific Changes Outperform Generic Guidance.} Fifteen of sixteen Components outperform their matching reminders in reliability, with at least one in every case across both harnesses and domains (Tables~\ref{tab:mechanism-factorization}--\ref{tab:generic-reminder-controls}). This motivates trace-grounded diagnoses that identify what to change and preserve.

\subsection{Mutate: Preserving Useful Behavior While Changing the Agent}
\label{sec:mutation-mechanisms}

\textbf{Mutation Quality Depends on How Changes Combine.} Six of eight Champions outperform either isolated Component in reliability, including all coding-task cases (Table~\ref{tab:mechanism-factorization}). Two general-task cases, one per harness, favor a subset. In one coding case, only the combination improves over Seed. These patterns motivate \emph{mutate} to preserve dependencies and check harmful interactions and useful complementarities in the complete Candidate. Appendix~\ref{app:rq3-mutate-details} reports interaction estimates.

\subsection{Gate: Deciding Which Changes to Retain}
\label{sec:gate-replay}

We compare original \emph{gate} rules with a constraint requiring higher mean success and incumbent-relative mean regression $\leq0.05$. A retained set $\mathcal{R}$ scores $1/|\mathcal{R}|$ if it contains the reporting-panel best Candidate, and zero otherwise. Scores average four method-specific pools per harness; reporting outcomes remain hidden from decisions.

\textbf{Admission Constraints Have Harness-Dependent Effects.} Constrained admission reduces best-Candidate retention from fifteen to six pools. Yet its size-normalized score falls for the source-editable harness and rises for the outer-layer-only harness on both domains (Table~\ref{tab:gate-constraint}). The increase reflects smaller sets retaining a best Seed alone. Admission quality therefore requires inspecting preservation, set size, and Candidate identity together. Appendix~\ref{app:rq3-gate-details} gives per-pool results.

\section{Recipes: From Empirical Findings to Operator Design}
\label{sec:design-insights}

Table~\ref{tab:design-hypotheses} connects the findings to operator design.

\begin{table}[htbp]
\caption{Cross-experiment findings and the design hypotheses they motivate.}
\label{tab:design-hypotheses}
\centering
\small
\begin{tabularx}{\textwidth}{@{}>{\raggedright\arraybackslash}m{0.22\textwidth}YY@{}}
\toprule
Integrated Finding & Supporting Evidence & Design Implication \\
\midrule
\textbf{Gains Involve Trade-offs.} & New coverage accompanies regressions and higher cost (Table~\ref{tab:overall}, Figures~\ref{fig:task-transitions}--\ref{fig:deployment-cost-curves}). & Consider gains, regressions, and cost in \emph{select} and \emph{gate}. \\
\midrule
\textbf{Gains Are Context-Dependent.} & Benefits vary with exposure and transfer target (Table~\ref{tab:ood-profile}). & Specify applicability in \emph{analyze} and \emph{mutate}. \\
\midrule
\textbf{Generation and Retention Shape Gains.} & Current scores, combinations, and admission can mislead search (Tables~\ref{tab:select-parent-quality}--\ref{tab:gate-constraint}). & Check subsequent gains, trace evidence, compatibility, and retained alternatives. \\
\bottomrule
\end{tabularx}
\par
\end{table}

\subsection{From Synthesis to Operator Revisions}
\label{sec:analysis-synthesis}
\label{sec:optimization-plan}

\begin{figure}[htbp]
\centering
\includegraphics[width=0.94\textwidth]{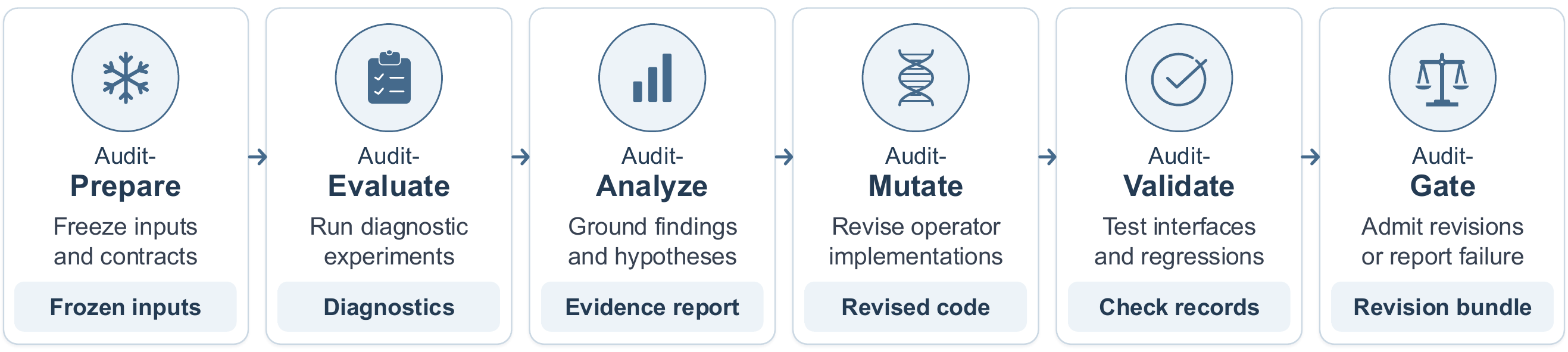}
\caption{EvoAudit diagnoses archived evolution runs and returns validated operator revisions.}
\label{fig:evoaudit-workflow}
\end{figure}

We propose \emph{EvoAudit}, an agent-assisted workflow that runs diagnostics on archived artifacts and returns operator revisions with evidence and validation records (Figure~\ref{fig:evoaudit-workflow}).

\label{sec:design-freeze}\label{sec:algorithm-construction}Installing its four revisions in A-Evolve gives \emph{Audit-Guided Evolution} (AGE). Other operators, model and evaluator contracts, and harness editing constraints remain fixed.

\subsection{AGE Implementation}
\label{sec:age-implementation}

\textbf{Recipe and Evidence Flow.} AGE installs four fixed operator revisions in the A-Evolve recipe described in Appendix~\ref{app:operator-implementations}, shared across both benchmarks and harnesses. Each generation selects a Parent, collects fresh Train trajectories, diagnoses behavior, and proposes one Candidate. Framework scope checks and canonical Validation evaluation precede admission and recording. The reference \emph{rollout} and \emph{record} implementations are unchanged, and the optional \emph{validate} stage is disabled. MiniSWE permits changes to inner source and outer artifacts, whereas Codex permits only outer-artifact changes. The evaluator, task partitions, model settings, resource limits, and evolution operators remain fixed. Online decisions use Train and Validation evidence.

\textbf{Run Configuration.} Each run uses 20 generations, one proposed Candidate per generation, and search random seed 0. The target harnesses are MiniSWE 2.4.5 and Codex 0.143.0, both using GPT-5.4 with high reasoning effort. Diagnosis and mutation use Codex 0.146.0 with GPT-5.4 and xhigh effort through Harbor. Every rollout executes the 50 Train tasks afresh. Canonical scoring executes each Validation task once, using 20 tasks on $\tau^3$-Banking and 19 on Terminal-Bench~2. Frozen Champions receive separate eight-pass reporting evaluations.

\textbf{Select: Regression-Aware Tie-Breaking.} AGE retains the highest-score objective and revises only exact ties. Let $\mathcal{P}$ be the eligible Parents with finite recorded Validation scores $s(C)$, and let $\mathcal{T}=\arg\max_{C\in\mathcal{P}}s(C)$. When comparable task-level evidence is available, selection minimizes Seed-relative regressions within $\mathcal{T}$:
\[
R(C)=\sum_{t\in\mathcal{D}_{\mathrm{val}}}\mathbb{I}[\bar r_t(C)<\bar r_t(A_0)],
\qquad
P\sim\operatorname{Uniform}\!\left(\arg\min_{C\in\mathcal{T}}R(C)\right),
\]
where $\mathcal{D}_{\mathrm{val}}$ is the Validation task set and $\bar r_t$ is the mean recorded reward on task $t$. Comparable evidence requires a unique eligible Seed evaluation, matching task-set hashes and task identities, equal per-task repetition counts, and completed trials with finite rewards for every tied Candidate and the Seed. Otherwise, the seeded generator samples uniformly from the original highest-score tie set. A unique maximizer is selected directly, and an empty eligible scored set stops selection. The rule reuses recorded outcomes and preserves the highest-score objective.

\textbf{Analyze: Behavior-Grounded Diagnosis.} The analyzer collects the current rollout and, when available, one preceding rollout along the selected lineage, retaining the last 30 records in their stored order. Each judge input contains chronological tool calls, successful and failed tool results, and the completion signal. Tool arguments and results are capped at 600 and 800 characters, respectively. Each rendered trajectory uses at most 12,000 characters, preserving approximately two-thirds from its beginning and one-third from its end when truncated. Interrupted executions can include a bounded termination message. The behavior bundle excludes benchmark rewards, verifier outputs, hidden tests, raw task text, and raw-case paths.

The judge returns \texttt{score}, \texttt{category}, \texttt{outcome}, and \texttt{failure\_reason}. Scores 0--3 indicate visible breakdown, 4--6 indicate partial or ambiguous progress, and 7--10 indicate visible completion supported by an appropriate check or comparably decisive result. Verdicts distinguish \texttt{Observed:} behavior from \texttt{Inference:} explanations and ground completion judgments in trajectory evidence. These behavioral scores guide diagnosis, while benchmark scores drive \emph{select} and \emph{gate}. Each request permits one attempt with a 600-second timeout. Invalid or failed responses return an unknown verdict with sentinel score $-1$.

The analyzer exports trajectories, behavioral signals, and verdicts in a structured evidence bundle. The selected text view uses at most 30,000 characters for $\tau^3$-Banking and 60,000 for Terminal-Bench~2, omitting trailing whole records beyond these budgets. Mutation consumes at most a 50,000-character prefix of this view. Both truncation steps add explicit omission markers.

\textbf{Mutate: Mechanism-Linked, Dependency-Closed Changes.} The mutation agent receives the selected behavior evidence, Parent workspace, prompt and skill inventory, and editing contract. Before choosing files, its instructions require the observed behavior, inferred mechanism, replacement behavior, activation conditions, and expected trade-off. A recurring concrete pattern or a single severe, visibly supported breakdown can motivate a change, while ambiguous evidence favors preserving current behavior. The agent must inspect affected interfaces and their producers and consumers, make one coherent intervention, preserve useful behavior in successful traces, and check for conflicting or duplicated guidance. Dependency-closed changes may span multiple files and layers.

The agent is instructed to edit the permitted target directly, inspect its final diff, run proportionate structural checks, and propose a validation case for the expected benefit. Outputs retain the patch, changed-file list, surface-check report, agent response, and usage record. Structural checks enforce editing surfaces and interfaces, while the mutation instructions specify evidence grounding, dependency closure, and behavior preservation. Execution errors stop the operator and are recorded with their diagnostic information.

\textbf{Gate: Score-First, Cost-Aware Admission.} AGE retains strict Parent-relative score improvement and adds a cost tie-break. For a Candidate $C$ and its matched Parent $P$, with finite Validation scores, admission is
\[
\operatorname{accept}(C,P)=\mathbb{I}\!\left[s(C)>s(P)\;\lor\;\bigl(s(C)=s(P)\land Q(C,P)\land c(C)<c(P)\bigr)\right].
\]
Here $c$ is the recorded USD cost of canonical Validation evaluation. The comparability predicate $Q$ requires matching task-set hashes, finite strictly positive costs, and a direct Parent evaluation. Equal-score admission requires a verified, comparable cost reduction. When cost comparability fails, only the score-improvement branch applies. A valid matched Parent and finite scores are required, and all other Candidates are rejected. The operator reuses canonical evaluations and preserves score improvement as its primary objective.

\Needspace{10\baselineskip}
\subsection{Human Evaluation of the Operator Revisions}
\label{sec:human-study}

\begin{wraptable}[7]{r}{0.44\textwidth}
\vspace{-\intextsep}
\centering
\setlength{\abovecaptionskip}{0pt}
\setlength{\belowcaptionskip}{6pt}
\caption{Expert review of four revisions.}
\label{tab:human-evaluation}
\footnotesize
\setlength{\tabcolsep}{3pt}
\begin{tabularx}{\linewidth}{@{}Y>{\centering\arraybackslash}m{1.45cm}@{}}
\toprule
Measure & Mean / 8 \\
\midrule
Direction agreement & $6/8$ \\
Evidence support & $6.25/8$ \\
Problem addressing & $6.5/8$ \\
Proposal endorsement & $6.75/8$ \\
\bottomrule
\end{tabularx}
\par
\end{wraptable}

Eight experts reviewed all four operator revisions. Each chose a direction before seeing the proposal, then judged its evidence, whether it addressed the stated problem, and whether it should enter validation. Table~\ref{tab:human-evaluation} reports the mean number of experts per revision. Overall, expert judgments broadly align with EvoAudit's proposed directions and support validating the resulting revisions. Appendix~\ref{app:operator-human-review} details the protocol.

\WFclear
\subsection{Comparison with Existing Evolution Methods}
\label{sec:algorithm-validation}

Table~\ref{tab:final-algorithm} compares AGE with the four methods across both benchmarks and harnesses.

\begin{table}[htbp]
\caption{Comparison of AGE with existing evolution methods. Score changes are relative to the Seed. Costs follow Equation~\ref{eq:cost}, with $k=8$ for deployment and 20 generations for evolution.}
\label{tab:final-algorithm}
\centering
\resizebox{0.9\textwidth}{!}{%
\begin{tabular}{lllcccccccc}
\toprule
\multirow{2}{*}{Benchmark} & \multirow{2}{*}{Harness} & \multirow{2}{*}{Variant} & \multicolumn{2}{c}{Reliability} & \multicolumn{2}{c}{Coverage} & \multicolumn{2}{c}{Task Turnover} & \multicolumn{2}{c}{Cost} \\
\cmidrule(lr){4-5}\cmidrule(lr){6-7}\cmidrule(lr){8-9}\cmidrule(lr){10-11}
& & & $\mathrm{Avg}@8$ & $\Delta\mathrm{Avg}$ & $\mathrm{Pass}@8$ & $\Delta\mathrm{Pass}$ & Gained & Lost & Deployment & Evolution \\
\midrule
\multirow{12}{*}{$\tau^3$-Banking} & \multirow{6}{*}{MiniSWE} & Seed & $27.4$ & \NA & $54.6$ & \NA & \NA & \NA & $6.2$ & \NA \\
 &  & AHE & $32.9$ & \textcolor{green!50!black}{$+5.4$} & \textbf{66.0} & \textcolor{green!50!black}{$+11.3$} & \textbf{15} & 4 & $7.1$ & $1235.0$ \\
 &  & HyperAgents & $32.6$ & \textcolor{green!50!black}{$+5.2$} & $57.7$ & \textcolor{green!50!black}{$+3.1$} & 13 & 10 & $6.6$ & $867.0$ \\
 &  & A-Evolve & $26.3$ & \textcolor{red!75!black}{$-1.2$} & $55.7$ & \textcolor{green!50!black}{$+1.0$} & 7 & 6 & $5.5$ & $1057.6$ \\
 &  & GEPA & $28.2$ & \textcolor{green!50!black}{$+0.8$} & $58.8$ & \textcolor{green!50!black}{$+4.1$} & 10 & 6 & $7.9$ & $2262.8$ \\
 &  & AGE & \textbf{34.0} & \textcolor{green!50!black}{$+6.6$} & \textbf{66.0} & \textcolor{green!50!black}{$+11.3$} & 13 & \textbf{2} & $6.1$ & $1070.6$ \\
\cmidrule(l){2-11}
 & \multirow{6}{*}{Codex} & Seed & $30.3$ & \NA & $56.7$ & \NA & \NA & \NA & $7.9$ & \NA \\
 &  & AHE & $36.2$ & \textcolor{green!50!black}{$+5.9$} & $62.9$ & \textcolor{green!50!black}{$+6.2$} & 12 & 6 & $8.9$ & $1385.1$ \\
 &  & HyperAgents & $36.9$ & \textcolor{green!50!black}{$+6.6$} & $66.0$ & \textcolor{green!50!black}{$+9.3$} & 11 & \textbf{2} & $8.8$ & $1175.8$ \\
 &  & A-Evolve & $28.1$ & \textcolor{red!75!black}{$-2.2$} & $57.7$ & \textcolor{green!50!black}{$+1.0$} & 9 & 8 & $7.6$ & $1523.6$ \\
 &  & GEPA & $29.5$ & \textcolor{red!75!black}{$-0.8$} & $53.6$ & \textcolor{red!75!black}{$-3.1$} & 6 & 9 & $7.8$ & $2372.8$ \\
 &  & AGE & \textbf{38.1} & \textcolor{green!50!black}{$+7.9$} & \textbf{67.0} & \textcolor{green!50!black}{$+10.3$} & \textbf{15} & 5 & $8.9$ & $1480.8$ \\
\midrule
\multirow{12}{*}{\shortstack{Terminal-\\Bench~2}} & \multirow{6}{*}{MiniSWE} & Seed & $65.9$ & \NA & $86.5$ & \NA & \NA & \NA & $6.2$ & \NA \\
 &  & AHE & $70.2$ & \textcolor{green!50!black}{$+4.4$} & $87.6$ & \textcolor{green!50!black}{$+1.1$} & 4 & 3 & $5.1$ & $853.4$ \\
 &  & HyperAgents & $67.7$ & \textcolor{green!50!black}{$+1.8$} & $85.4$ & \textcolor{red!75!black}{$-1.1$} & 3 & 4 & $5.3$ & $723.4$ \\
 &  & A-Evolve & $68.0$ & \textcolor{green!50!black}{$+2.1$} & $86.5$ & $0.0$ & 2 & \textbf{2} & $6.3$ & $983.9$ \\
 &  & GEPA & $68.5$ & \textcolor{green!50!black}{$+2.7$} & $88.8$ & \textcolor{green!50!black}{$+2.2$} & 4 & \textbf{2} & $5.6$ & $1629.0$ \\
 &  & AGE & \textbf{71.8} & \textcolor{green!50!black}{$+5.9$} & \textbf{89.9} & \textcolor{green!50!black}{$+3.4$} & \textbf{5} & \textbf{2} & $5.2$ & $982.7$ \\
\cmidrule(l){2-11}
 & \multirow{6}{*}{Codex} & Seed & $69.9$ & \NA & $89.9$ & \NA & \NA & \NA & $7.1$ & \NA \\
 &  & AHE & $72.1$ & \textcolor{green!50!black}{$+2.1$} & $91.0$ & \textcolor{green!50!black}{$+1.1$} & \textbf{5} & 4 & $7.8$ & $1065.5$ \\
 &  & HyperAgents & $72.6$ & \textcolor{green!50!black}{$+2.7$} & $86.5$ & \textcolor{red!75!black}{$-3.4$} & 2 & 5 & $8.9$ & $987.2$ \\
 &  & A-Evolve & $68.0$ & \textcolor{red!75!black}{$-2.0$} & $87.6$ & \textcolor{red!75!black}{$-2.2$} & 1 & 3 & $8.2$ & $1254.1$ \\
 &  & GEPA & $68.1$ & \textcolor{red!75!black}{$-1.8$} & $86.5$ & \textcolor{red!75!black}{$-3.4$} & 2 & 5 & $8.1$ & $1939.2$ \\
 &  & AGE & \textbf{73.5} & \textcolor{green!50!black}{$+3.5$} & \textbf{92.1} & \textcolor{green!50!black}{$+2.2$} & 3 & \textbf{1} & $6.8$ & $1160.5$ \\
\bottomrule
\end{tabular}}
\par
\end{table}

\textbf{Combined Revisions Improve Outcomes Across Settings.} AGE improves reliability and coverage over A-Evolve in all four settings and loses fewer Seed-covered tasks in three, tying in the fourth. This consistency across domains and harnesses supports the practical value of evidence-guided revision.

\section{Discussion}
\label{sec:discussion}

\textbf{Feedback Can Be Narrower Than a Change's Effects.} Harness changes affect many behaviors, while evaluation samples finite executions and compresses their trade-offs. A prompt, skill, or execution routine can help the tasks that supply feedback while disrupting behavior elsewhere. This mismatch offers a common explanation for task regressions, context-dependent change interactions, and selective transfer. It suggests that practical evolution depends on how well feedback represents the consequences of a change, including the resources needed to deploy it.

\textbf{Evidence Must Match the Decision.} Current performance, improvement potential, change compatibility, and retention are distinct objectives. A high-scoring Parent need not yield the greatest subsequent gain, a useful Component can interfere with other changes, and a conservative gate can discard alternatives. Together, the diagnostics show why evidence supporting one decision need not support the next. Evaluation should test the desired consequence of each decision, rather than reuse the same aggregate score throughout evolution.

\textbf{Toward Decision-Specific Feedback.} Evaluate \emph{select} through subsequent gains, ground \emph{analyze} and \emph{mutate} in traces and dependencies, and assess \emph{gate} through both retained quality and preserved alternatives. Deployment cost and external evaluation then determine how broadly these benefits remain useful. EvoAudit and AGE illustrate how these complementary checks can guide a coordinated redesign while preserving the reference method's execution and evaluation contracts.

\section{Conclusion}

Our study shows that harness evolution redistributes observed task success, bringing both gains and regressions. Across harnesses, task domains, and evolution paradigms, practical improvement depends on productive Parents, compatible changes, preservation through search, and generalization at affordable cost. EvoAudit connects these findings to operator revisions, and AGE improves reliability and coverage over its reference method in all four settings. These results motivate evaluating evolution by the useful behavior it creates and sustains, beyond final benchmark scores.

\subsection*{Ethics Statement}
Experiments run in isolated benchmark environments rather than live user-facing services. Evolving agent instructions and code can introduce regressions and unintended tool use. We retain fixed evaluation contracts and harness-specific editing permissions, and examine regressions and transfer alongside aggregate performance. Participants in the human evaluation provided informed consent. The expert-review protocol evaluates the evidence, reasoning, and proposed approaches for operator revisions, as described in Appendix~\ref{app:operator-human-review}. Neither benchmark gains nor expert endorsement establishes safety for deployment.

\FloatBarrier
\bibliographystyle{plainnat}
\bibliography{references}

\clearpage
\appendix

\section{Operator Implementations of the Compared Methods}
\label{app:operator-implementations}

We describe the implementations used in our experiments through the seven operator stages in Table~\ref{tab:method-operators}. All four methods share fixed target-model and evaluator contracts. MiniSWE permits dependency-closed changes to inner source code as well as prompts, skills, configuration, and orchestration; Codex permits only changes outside its immutable executable and source. HyperAgents additionally permits changes to designated evolution operators.

\textbf{Evolution Paradigms.} Our design-based grouping distinguishes implementation-centric recipes, AHE and HyperAgents, from reflection-centric recipes, A-Evolve and GEPA. The former organize revision around engineering the agent implementation or improvement machinery; the latter organize it around distilling execution experience and feedback. Both use evidence and may modify code within the harness's legal surface. The grouping follows recipe design, while the harness determines the permitted editing surface. Codex permits changes to external orchestration code while keeping its executable and inner source immutable.

\begin{table}[htbp]
\caption{Initial operator policies of the four methods. Dashes indicate an omitted optional stage. HyperAgents can subsequently revise its permitted operators.}
\label{tab:method-operators}
\centering
\small
\begin{tabularx}{\textwidth}{lYYYY}
\toprule
Operator & AHE & HyperAgents & A-Evolve & GEPA \\
\midrule
select & Newest eligible Parent & Score-weighted sampling with a child-count penalty & Highest-scoring eligible Parent & Per-task Pareto-front sampling \\
\midrule
rollout & Reuse Parent evaluation traces & Reuse Parent evaluation traces & Fresh Train trajectories & Fresh Train trajectories \\
\midrule
analyze & Per-task debugger reports & Metrics and browsable raw traces & Behavior-only trajectory judge & Focused reflective examples \\
\midrule
mutate & One testable target change & Edit the target and allowed operators & Distill experience into target changes & Reflection-guided target revision \\
\midrule
validate & \NA & Python syntax checks & \NA & Strict improvement on the Parent's Train batch \\
\midrule
gate & Admit valid completed evaluations & Admit valid completed evaluations & Strict improvement over the matched Parent & Admit valid completed evaluations \\
\midrule
record & Outcomes and change evidence & Lineage experience & Outcomes and rationale & Reflection and batch experience \\
\bottomrule
\end{tabularx}
\par
\end{table}

Search scores use Train tasks for AHE and HyperAgents, and Validation tasks for A-Evolve and GEPA. The \emph{validate} operator checks proposals, while the Validation split supplies evaluation tasks. Shared scope and evaluation-eligibility checks apply to all four methods, including those that omit the optional \emph{validate} stage. Test outcomes remain excluded from search.

\subsection{AHE}

\textbf{Selection and Diagnosis.} AHE emphasizes observability-grounded revision~\citep{lin2026ahe}. Its \emph{select} operator chooses the newest eligible Parent, not necessarily the highest-scoring one. The \emph{rollout} operator reuses that Parent's evaluation traces. The \emph{analyze} operator produces task-level debugger reports: failure reports identify the failure point, possible cause, and corrective lesson, while successful traces expose reusable strategies and fragile behavior. Mixed outcomes on a task can therefore inform the diagnosis alongside outright failures. An overview links these reports to prior changes and their observed effects.

\textbf{Revision and Admission.} The mutation agent is instructed to make one coherent, testable change, supported by task evidence and an identified execution path. It considers whether to retain, revise, or roll back the preceding change and pursue a different explanation. A change manifest records the proposed mechanism, affected files, predicted fixes, and behaviors at risk. Debugger reports supply hypotheses for revision. There is no separate optional \emph{validate} stage. The \emph{gate} operator admits a completed, eligible evaluation without requiring a higher score, so the next Parent can come from a lower-scoring revision. The \emph{record} operator preserves outcomes and change evidence. Admission and continuation are thus distinct from final Champion reporting.

\subsection{HyperAgents}

\textbf{Selection and Evidence.} HyperAgents permits revision of the improvement procedure as well as the target agent~\citep{zhang2026hyperagents}. Its initial \emph{select} policy samples eligible Parents using a score-dependent weight and a penalty for having many children. Higher scores increase sampling weight, while the child-count penalty reduces repeated expansion of the same Parent. The \emph{rollout} operator reuses Parent evaluation traces. The initial \emph{analyze} operator supplies metrics and browsable raw traces, leaving their interpretation largely to the mutation agent rather than first producing AHE-style debugger reports.

\textbf{Revision and Admission.} Mutation may change the target and permitted operator implementations while leaving the evaluator and other fixed contracts unchanged. The mutation instructions favor substantive target changes but allow an operator-only revision when the evidence supports a process-level change. An operator revision takes effect at its next invocation, updating the initial policies in Table~\ref{tab:method-operators} for subsequent Generations. The initial \emph{validate} stage checks Python syntax in target files and the mutation operator. The initial \emph{gate} admits completed, eligible evaluations without a Parent-relative score threshold. The \emph{record} operator stores lineage experience, including the Parent, outcome, and score. This editable search procedure distinguishes HyperAgents from target-only revision under fixed operators.

\subsection{A-Evolve}

\textbf{Selection and Behavioral Feedback.} Our A-Evolve implementation operationalizes experience-driven adaptation~\citep{lin2026aevolve}. The \emph{select} operator chooses the highest-scoring eligible Parent, and \emph{rollout} collects fresh Train trajectories. The \emph{analyze} operator compresses observable behavior, including tool use, repeated actions, errors, and completion signals, for a trajectory-only judge. This judge evaluates behavior without benchmark reward labels. Its proxy judgments provide feedback for revision, rather than replacing the benchmark score used for admission.

\textbf{Revision and Admission.} The \emph{mutate} operator distills these observations into reusable task-solving guidance and executable target changes. Its instructions emphasize recurrent failure patterns, self-contained skills with their supporting dependencies, and preservation of unrelated behavior. Prompt and skill settings organize revision within the Harness's legal editing boundary. The optional \emph{validate} stage is disabled. After canonical evaluation, the strict \emph{gate} admits a Candidate only if its Validation score exceeds that of the matched Parent. The \emph{record} operator retains outcomes and the revision rationale. Unlike AHE and the initial HyperAgents policy, A-Evolve combines greedy Parent choice with score-improving admission.

\subsection{GEPA}

\textbf{Selection and Reflection.} GEPA combines trajectory-based reflection with per-task Pareto selection~\citep{agrawal2026gepa}. The \emph{select} operator identifies eligible Candidates attaining the best observed score on at least one task, prunes redundant coverage, and samples according to the remaining per-task memberships. This can preserve access to task specialists without selecting solely by aggregate score. The \emph{rollout} operator collects fresh Train trajectories, and \emph{analyze} assembles task examples, outputs, and feedback for focused reflection.

\textbf{Revision Scope.} The named reflection units differ by benchmark. On $\tau^3$-Banking, MiniSWE focuses on its system-prompt configuration, while Codex alternates between its prompt and task-execution skill. On Terminal-Bench~2, both harnesses use the entire target as the reflection unit. These units guide evidence organization rather than impose text-only filesystem permissions. The frozen Champion from MiniSWE with GEPA on Terminal-Bench~2 retains inner-source changes to task-conditioned guidance and shell execution. Reflection-centric therefore describes the revision process, not the absence of code changes. These reflection units are distinct from the post-hoc Components isolated in our diagnostics.

\textbf{Validation and Admission.} The \emph{validate} operator evaluates the proposal on the same Train task batch used for the Parent rollout and requires a complete, scorable comparison with strict improvement. This feedback-informed proposal filter precedes canonical Validation evaluation. The subsequent \emph{gate} checks completion and eligibility without imposing a second Parent-relative improvement threshold. The \emph{record} operator preserves reflection choices, proposals, and batch comparisons. Thus, GEPA's improvement filter lies in \emph{validate}, whereas A-Evolve's lies in \emph{gate}, and they use different task evidence.

\textbf{Implications for the Analysis.} These implementations differ in how they allocate search, summarize evidence, couple target and operator changes, and filter proposals. Similar final gains can therefore arise through different processes. These differences motivate diagnostics of selection, diagnosis, change compatibility, and admission.

\section{Phenomenology: Additional Protocols and Analyses}
\label{app:rq1-details}

This appendix supplements the Phenomenology analysis in Section~\ref{sec:main-results} with case-level comparisons, task-transition stability, deployment-cost accounting, and split-level heatmaps. Section~\ref{app:deployment-cost} provides the numerical cost table, and Section~\ref{app:split-transition-heatmaps} contains the split-level heatmaps.

\subsection{Matched Outcome Comparisons}
\label{app:rq1-outcomes}

\textbf{Evaluation and Interpretation.} Every configuration in Table~\ref{tab:overall} uses the same benchmark tasks and eight complete passes. Each of the eight Champions per benchmark is paired with its harness Seed. The completed comparison contains 7,760 selected $\tau^3$-Banking executions and 7,120 Terminal-Bench~2 executions, including the two Seeds once per benchmark. Each method--harness--benchmark combination contributes one canonical evolution lineage.

\textbf{Metric Agreement and Disagreement.} Across the sixteen Champions, eleven improve reliability and ten improve coverage. Eight improve both, four move the metrics in opposite directions, one improves reliability with unchanged coverage, and three reduce both. The opposite-sign cases occur in reflection-centric general-task evolution and implementation-centric coding-task evolution, each with both harness types. Thus, the distinction between consistency and reach is not confined to one harness type, task domain, or paradigm, even though agreement is more common than disagreement.

\textbf{Interpreting Trends Across Dimensions.} Reliability improves in all eight implementation-centric lineages across harness types and task domains. In the reflection-centric group, it improves in three of four source-editable cases but none of the four outer-layer-only cases. The paradigm-level pattern is therefore harness-dependent. Recipes also differ in feedback, search, and admission design, suggesting concrete mechanisms for further study.

\subsection{Task-Level Turnover and Pass Stability}
\label{app:rq1-transitions}

\textbf{Transition Construction.} Each $9\times9$ heatmap pairs a task's Seed and Champion success counts from zero through eight, using complete $97\times8$ evaluations on $\tau^3$-Banking or $89\times8$ evaluations on Terminal-Bench~2. Panels within each benchmark use the same task set and color scale. The split heatmaps retain the original Train, Validation, and Test task identities.

\textbf{Coverage Turnover Across All Settings.} All sixteen Champions both gain and lose covered tasks. Gains range from one to fifteen tasks and losses from two to ten, with both types present in every harness--task--paradigm combination. A positive net coverage delta therefore represents the balance of newly covered and lost tasks rather than preservation of the Seed's solved set. This shared pattern motivates paired gain/loss accounting without requiring a specific method to have the largest net expansion.

\textbf{Within-Task Reliability Also Redistributes.} Every panel contains tasks on both sides of the success-count diagonal. Evolution can improve repeated success on already covered tasks, reduce it elsewhere, and change which tasks are covered at all. The shared combination of these movements explains why neither a net coverage gain nor a reliability gain summarizes preservation.

\textbf{Pass-Half Stability.} We compare passes 1--4 with passes 5--8 to assess the stability of aggregate trends and task transitions. Some aggregate directions differ between halves on $\tau^3$-Banking, showing sensitivity to execution variability.

\subsection{Behavioral Audit of Task-Level Transitions}
\label{app:rq1-behavior}

\textbf{Case Selection and Evidence.} We identify the largest positive and negative changes in eight-pass success count within each lineage, breaking ties lexicographically by task identifier before trace inspection. The reported analysis contains 22 task--lineage comparisons with trace-linked behavioral evidence. For each increase, we inspect the first failed Seed execution and the first successful Champion execution in canonical pass order; for each decrease, we inspect the first successful Seed execution and the first failed Champion execution. The audit uses archived actions, observations, submitted artifacts, and verifier records.

\textbf{Behavior Categories.} We assign one primary category to each reference pair based on the concrete behavioral difference linked to its task outcome. The summary in Table~\ref{tab:behavior-audit-summary} appears in the main text; Table~\ref{tab:behavior-audit-cases} illustrates the underlying trace evidence. The agent-assisted audit examines execution behavior; the human evaluation reviews operator revisions.

\textbf{What Changes Behind a Flip.} Three themes recur across the selected cases with both harness types, task domains, and evolution paradigms. First, state and dependency management affects the complete workflow. Successful general-task executions order operations so that one preserves the conditions needed by another, while successful coding executions preserve source data before inspection and deliver self-contained artifacts. Second, exact grounding remains fragile after the main workflow succeeds: filed requests can contain an incorrect date, and reconstructed text can contain a single wrong character. Third, intermediate success can fall short of task completion. An agent can open one requested account and stop too early, announce a transfer without completing it, or pass its own program checks while leaving an unacceptable final artifact. These observations motivate checking preserved state, exact arguments, and the final execution boundary alongside the headline success rate. Tool recovery and implementation debugging provide complementary case evidence.

\begin{table}[htbp]
\centering
\caption{Illustrative trace contrasts on $\tau^3$-Banking and Terminal-Bench~2. Counts are Seed $\rightarrow$ Champion successes out of eight; paired Champion counts follow the listed method order.}
\label{tab:behavior-audit-cases}
\small
\begin{tabularx}{\textwidth}{@{}>{\raggedright\arraybackslash}m{0.11\textwidth}>{\raggedright\arraybackslash}m{0.23\textwidth}cY@{}}
\toprule
Domain & Task and lineage & Count & Observed difference in the reference executions \\
\midrule
General & Task 054, MiniSWE with A-Evolve & $1\to4$ & Seed handles replacement and dispute first, blocking the requested credit-limit increase. Champion completes the increase before those dependent actions. \\
\addlinespace
General & Task 037, MiniSWE with A-Evolve and GEPA & $7\to3/1$ & Both Champions successfully file disputes but use the wrong issue-noticed date. Seed's successful reference uses the date required by the evaluator. \\
\addlinespace
General & Task 056, Codex with AHE & $5\to0$ & Champion terminates after opening checking. Seed continues the interaction and completes the additional savings and funding requests. \\
\addlinespace
General & Task 090, Codex with GEPA & $1\to5$ & Seed repeatedly fails through a generic tool wrapper. Champion switches to the available direct lookup and completes the investigation. \\
\midrule
Coding & \texttt{db-wal-recovery}, Codex with GEPA & $0\to7$ & Seed loses the original write-ahead log during inspection and reconstructs incorrect values. Champion preserves a copy before repair and verifies the recovered values. \\
\addlinespace
Coding & \texttt{polyglot-c-py}, MiniSWE with HyperAgents & $1\to8$ & Seed leaves test binaries and a script beside the required single source file. Champion leaves only the required file; the contrast concerns final directory contents. \\
\addlinespace
Coding & \texttt{path-tracing}, MiniSWE with A-Evolve & $2\to6$ & Seed's wrapper relies on a workspace binary unavailable in the evaluation jail. Champion submits a self-contained renderer. \\
\addlinespace
Coding & \texttt{path-tracing}, Codex with GEPA & $6\to1$ & The direction reverses: Champion replaces its renderer with a workspace-dependent wrapper that matches the image locally but fails in the jail. \\
\addlinespace
Coding & \texttt{gcode-to-text}, Codex with HyperAgents and A-Evolve & $8\to4/5$ & Champions recover the text geometry but confuse \texttt{o} with \texttt{0}, or \texttt{1} with \texttt{i}. Checking the written file does not correct the transcription. \\
\bottomrule
\end{tabularx}
\end{table}

\subsection{Deployment-Cost Accounting and Comparisons}
\label{app:rq1-cost-details}

\textbf{A Shared Execution Ledger.} Each configuration has 776 selected logical trials on $\tau^3$-Banking or 712 on Terminal-Bench~2. Coverage and cost use this ledger and the metrics in Section~\ref{sec:prelim-metrics}. Prefix curves follow the canonical pass order $1,\ldots,k$. Superseded attempts remain in the audit ledger but are excluded from deployment accounting.

\textbf{A Frequent Cost Premium, With Structured Exceptions.} Eight of the eleven reliability-improving Champions have higher mean deployment cost than Seed. Such premiums occur with both harness types, task domains, and paradigms. All five reliability-improving general-task Champions cost more, whereas three of the six reliability-improving coding-task Champions cost less. The latter comprise two implementation-centric cases and one reflection-centric case, all with the source-editable harness. This harness--task cluster shows how the resource trade-off varies by setting, motivating matched-resource comparisons alongside reliability.

\textbf{Cost and Performance Need Joint Interpretation.} All four outer-layer-only Champions on coding tasks cost more than Seed, although only two improve reliability. Coverage--cost curves relate empirical reach to deployment resources at each pass prefix. Evolution cost complements these curves by accounting for the resources used to obtain the evolved agent.

\subsubsection{Coverage and Deployment Cost: Detailed Results}
\label{app:deployment-cost}

Table~\ref{tab:deployment-cost} provides the numerical results behind Figure~\ref{fig:deployment-cost-curves}, using the metrics in Section~\ref{sec:prelim-metrics} and the ledger described in Appendix~\ref{app:rq1-cost-details}.

\begin{table}[htbp]
\caption{Deployment-side coverage and recorded costs on both primary benchmarks. Scores are percentages and costs are in USD.}
\label{tab:deployment-cost}
\centering
\resizebox{0.98\textwidth}{!}{%
\begin{tabular}{lllrrrrrr}
\toprule
Benchmark & Harness & Variant & $\mathrm{Pass}@1$ & $\mathrm{Pass}@2$ & $\mathrm{Pass}@4$ & $\mathrm{Pass}@8$ & Mean cost/trial & Recorded cost \\
\midrule
\multirow{10}{*}{$\tau^3$-Banking} & \multirow{5}{*}{MiniSWE}
& Seed         & $27.8$ & $39.2$ & $46.4$ & $54.6$ & $\$0.779$ & $\$604.52$ \\
& & AHE          & $28.9$ & $46.4$ & $56.7$ & $66.0$ & $\$0.891$ & $\$691.57$ \\
& & HyperAgents  & $33.0$ & $39.2$ & $50.5$ & $57.7$ & $\$0.819$ & $\$635.56$ \\
& & A-Evolve     & $26.8$ & $36.1$ & $48.5$ & $55.7$ & $\$0.690$ & $\$535.57$ \\
& & GEPA         & $29.9$ & $40.2$ & $51.5$ & $58.8$ & $\$0.985$ & $\$764.57$ \\
\cmidrule(l){2-9}
& \multirow{5}{*}{Codex}
& Seed         & $24.7$ & $42.3$ & $50.5$ & $56.7$ & $\$0.983$ & $\$762.45$ \\
& & AHE          & $34.0$ & $46.4$ & $53.6$ & $62.9$ & $\$1.108$ & $\$859.82$ \\
& & HyperAgents  & $36.1$ & $46.4$ & $60.8$ & $66.0$ & $\$1.104$ & $\$856.43$ \\
& & A-Evolve     & $28.9$ & $39.2$ & $49.5$ & $57.7$ & $\$0.955$ & $\$740.79$ \\
& & GEPA         & $33.0$ & $42.3$ & $49.5$ & $53.6$ & $\$0.978$ & $\$758.73$ \\
\midrule
\multirow{10}{*}{Terminal-Bench~2} & \multirow{5}{*}{MiniSWE} & Seed & $56.2$ & $74.2$ & $82.0$ & $86.5$ & $\$0.774$ & $\$551.17$ \\
 &  & AHE & $74.2$ & $77.5$ & $85.4$ & $87.6$ & $\$0.639$ & $\$455.27$ \\
 &  & HyperAgents & $64.0$ & $75.3$ & $79.8$ & $85.4$ & $\$0.667$ & $\$474.72$ \\
 &  & A-Evolve & $69.7$ & $77.5$ & $82.0$ & $86.5$ & $\$0.783$ & $\$557.29$ \\
 &  & GEPA & $68.5$ & $75.3$ & $86.5$ & $88.8$ & $\$0.701$ & $\$499.14$ \\
\cmidrule(l){2-9}
 & \multirow{5}{*}{Codex} & Seed & $69.7$ & $80.9$ & $87.6$ & $89.9$ & $\$0.887$ & $\$631.85$ \\
 &  & AHE & $73.0$ & $84.3$ & $87.6$ & $91.0$ & $\$0.971$ & $\$689.18$ \\
 &  & HyperAgents & $71.9$ & $84.3$ & $85.4$ & $86.5$ & $\$1.107$ & $\$785.63$ \\
 &  & A-Evolve & $69.7$ & $78.7$ & $84.3$ & $87.6$ & $\$1.031$ & $\$734.06$ \\
 &  & GEPA & $58.4$ & $76.4$ & $82.0$ & $86.5$ & $\$1.007$ & $\$716.64$ \\
\bottomrule
\end{tabular}}
\par
\end{table}

\subsection{Split-Level Task-Transition Heatmaps}
\label{app:split-transition-heatmaps}

\begin{figure}[!htbp]
\textbf{Train Split}\par\medskip
\centering
\includegraphics[width=0.7\textwidth]{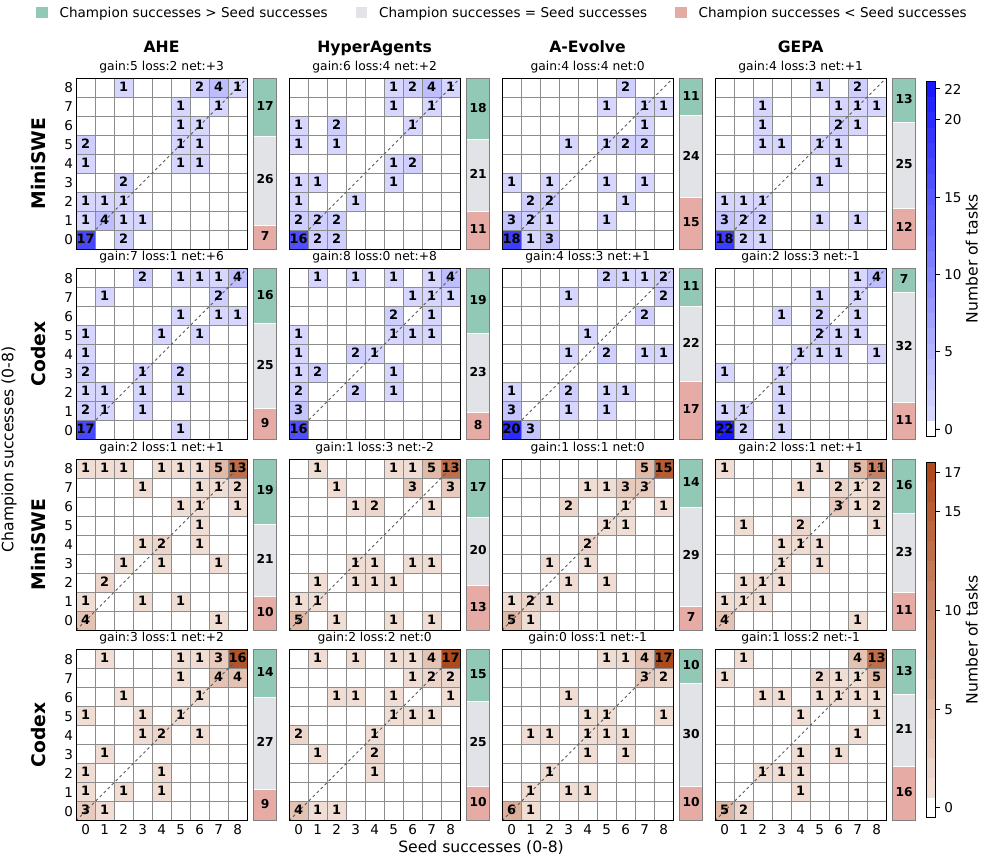}
\caption{Task-level Seed-to-Champion transitions over eight passes on the Train split. Each panel contains 50 $\tau^3$-Banking tasks (blue) or 50 Terminal-Bench~2 tasks (orange).}
\label{fig:split-train}
\label{fig:split-train-tau3}
\label{fig:split-train-tb2}
\end{figure}

\begin{figure}[!htbp]
\textbf{Validation Split}\par\medskip
\centering
\includegraphics[width=0.7\textwidth]{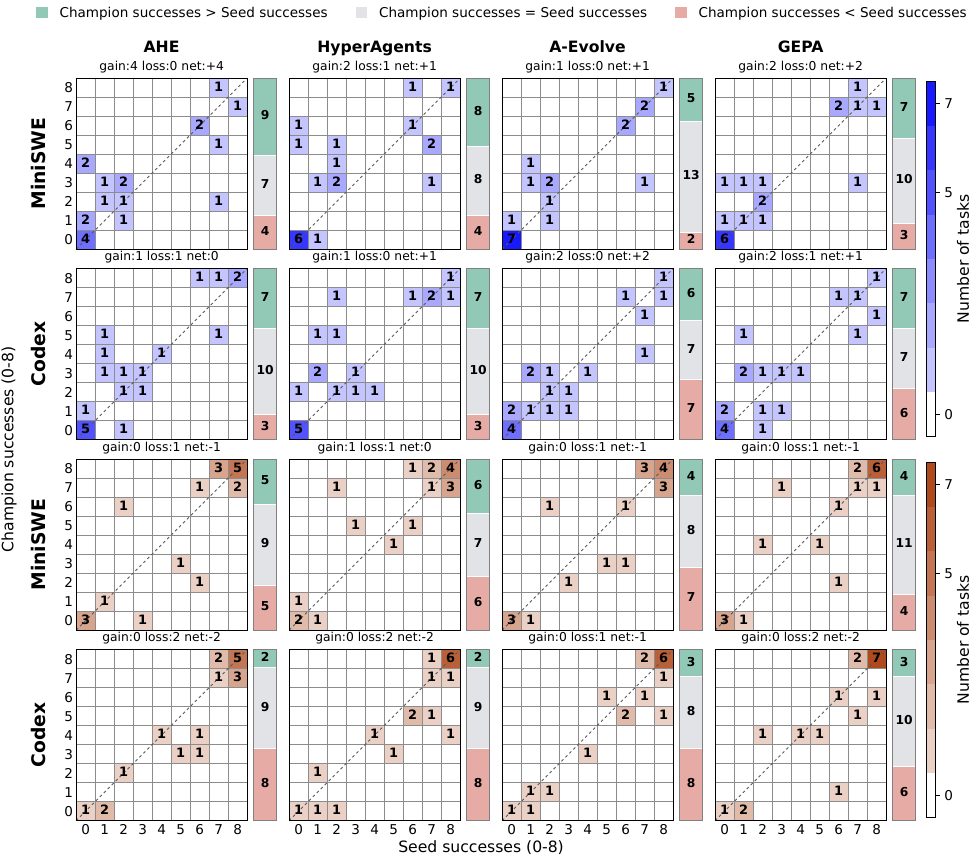}
\caption{Task-level Seed-to-Champion transitions over eight passes on the Validation split. Each panel contains 20 $\tau^3$-Banking tasks (blue) or 19 Terminal-Bench~2 tasks (orange).}
\label{fig:split-gate}
\label{fig:split-gate-tau3}
\label{fig:split-gate-tb2}
\end{figure}

\begin{figure}[!htbp]
\textbf{Test Split}\par\medskip
\centering
\includegraphics[width=0.7\textwidth]{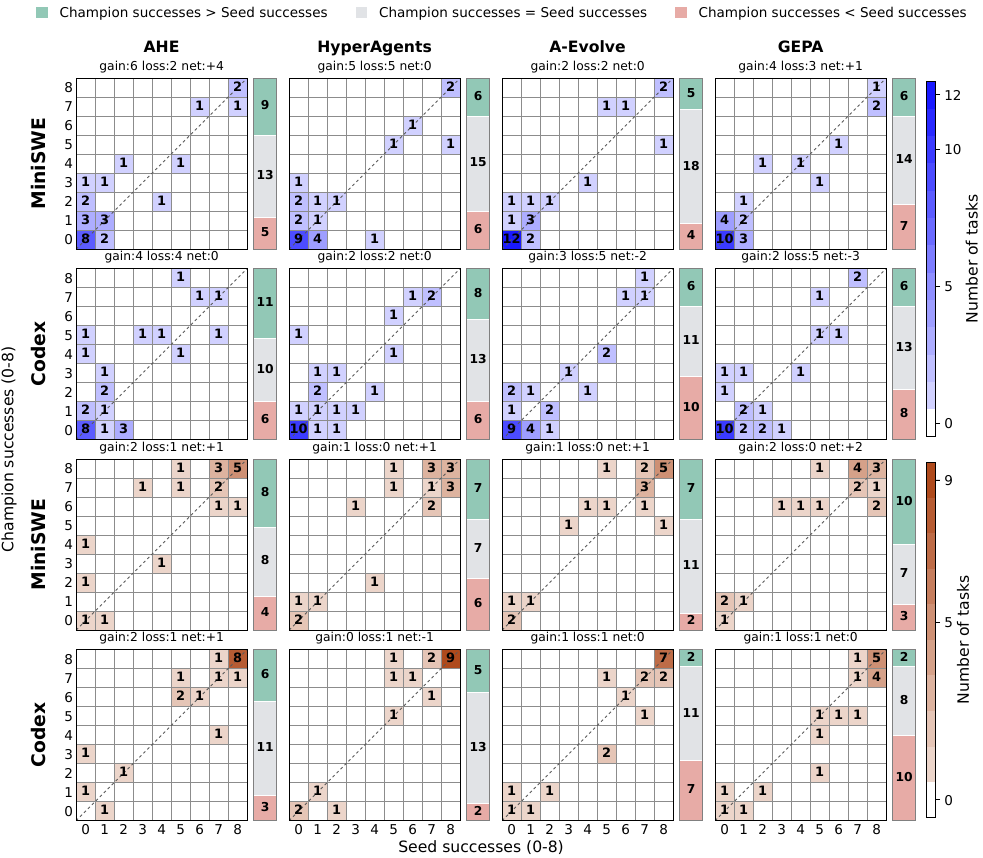}
\caption{Task-level Seed-to-Champion transitions over eight passes on the Test split. Each panel contains 27 $\tau^3$-Banking tasks (blue) or 20 Terminal-Bench~2 tasks (orange).}
\label{fig:split-sealed}
\label{fig:split-sealed-tau3}
\label{fig:split-sealed-tb2}
\end{figure}

Figures~\ref{fig:split-train}--\ref{fig:split-sealed} show eight-pass transitions by split, with $\tau^3$-Banking above Terminal-Bench~2.

\section{Generalization: Additional Protocols and Analyses}
\label{app:rq2-details}

This appendix supplements the Generalization analysis in Section~\ref{sec:transfer}. It separates task exposure during evolution from cross-benchmark transfer, specifies both external evaluation contracts, and collects absolute transfer scores in one table. The two primary benchmarks remain $\tau^3$-Banking and Terminal-Bench~2.

\subsection{Search Exposure and Split Accounting}
\label{app:rq2-exposure}

\textbf{Task Exposure and External Transfer.} Train trajectories supply evolution feedback, and Validation evidence can influence admission. Test tasks retain the source benchmark's domain and evaluation contract while remaining excluded from mutation feedback and admission decisions. External targets change the benchmark within the corresponding general-task or coding-task domain. The comparison follows these roles to examine gains beyond direct feedback and across benchmarks.

\textbf{Reuse and Denominators.} Table~\ref{tab:ood-profile} reuses the complete eight-pass measurements reported in Section~\ref{sec:main-results}. Split labels retain their roles during evolution. For a split, the change in successful executions is divided by $8N$ and the change in covered tasks by $N$, where $N$ is the full source-benchmark task count. The three split contributions therefore sum to the overall gain before rounding. External deltas use each target's own task set and matching Seed.

\textbf{Persistence Across Splits and Harnesses.} Ten of sixteen Champions have positive Test reliability contributions. Seven of eight implementation-centric cases improve across both harness types and task domains, with one source-editable general-task exception at $-0.1$ points. Reflection-centric cases improve in three of four source-editable settings but none of the four outer-layer-only settings. Thus, most implementation-centric gains reach feedback-excluded tasks, while reflection-centric persistence is more harness-dependent. On coding tasks, all outer-layer-only Champions lose Validation reliability, yet the implementation-centric Champions gain Test reliability. Gains therefore vary across splits as well as harness--paradigm combinations.

\subsection{General-Task Transfer Protocol}
\label{app:rq2-customer-transfer}

\textbf{Full-Benchmark Evaluation.} The general-task transfer comparison evaluates eight frozen $\tau^3$-Banking Champions and two harness Seeds on the full $\tau^2$-bench \citep{barres2025tau2}, covering retail, airline, and telecom. Following official benchmark-level aggregation, we compute each metric within each subdomain and take the equal-weight mean across the three subdomains for both Seed and Champion. All 278 tasks have eight valid outcomes per configuration, totaling 22,240 verified evaluations. These measurements supply the absolute scores in Table~\ref{tab:transfer-details} and matched-Seed gains in Table~\ref{tab:ood-profile}; deltas are computed before rounding. Our $\mathrm{Pass}@8$ measures at least one success in eight attempts, while the benchmark's $\mathrm{pass}^{8}$ measures consistency across all eight attempts.

\textbf{Neutral Adaptation and Quality Checks.} Within each harness, frozen neutral adapters translate message and tool protocols while preserving target-native user interaction, permissions, state transitions, and verification. All versions share these adapters, with Champion instructions retained unchanged. The evaluation contract fixes task identities, model settings, inference limits, and a score-independent recovery policy. Compatibility checks and smoke tests precede the full evaluation; reporting requires complete valid outcomes under the same contract.

\textbf{Coverage Contraction With a Metric-Specific Exception.} Seven of eight Champions lose coverage on the general-task transfer target. The sole positive case is reflection-centric and source-editable, at $+1.5$ points. The recurring contraction spans both harness types and paradigms, while reliability follows a different pattern: all four source-editable Champions improve, and three of four outer-layer-only Champions decline. Thus, more consistent success on the transfer target can coexist with narrower empirical reach, and coverage contraction should not be described as a uniform loss of all benefits.

\subsection{Coding-Task Transfer Protocol}
\label{app:rq2-software-transfer}

\textbf{Frozen Evaluation.} DeepSWE~v1.1 \citep{huang2026deepswe} tests transfer from terminal-computing tasks to long-horizon repository-level implementation without target-specific mutation or Champion-specific adaptation. The complete comparison contains four frozen Champions and the matching Seed for each harness, each with 113 official tasks and eight passes, or 904 selected trials per configuration. All 9,040 logical outcomes are complete and unique. Task identities, evaluator and model contracts, inference limits, and recovery rules remain fixed across these versions. Table~\ref{tab:transfer-details} reports the complete scores for both harnesses.

\textbf{A Target-Dependent Reversal Across Paradigms.} Seven of eight Champions gain coding-transfer coverage, compared with seven losing coverage on the general-task target. The coding gains span both harness types and paradigms, with one reflection-centric, source-editable exception at $-2.7$ points. All four source-editable Champions nevertheless lose coding-transfer reliability, by $0.2$--$2.3$ points, while all four improve it on the general-task target. Outer-layer-only coding-transfer reliability improves in two cases and declines in two. These results separate reach from consistency and show that transfer need not uniformly erode gains.

\subsection{Absolute Transfer Scores}
\label{app:transfer-scores}

Table~\ref{tab:transfer-details} collects the absolute scores for both transfer targets. Each method row denotes its frozen Champion under the specified harness, Seed is the matched unevolved agent.

\begin{table}[htbp]
\caption{Absolute performance on external transfer benchmarks. Scores are percentages. $\tau^2$-bench evaluates agents evolved on $\tau^3$-Banking using the official equal-weight aggregation across its three domains; DeepSWE~v1.1 evaluates agents evolved on Terminal-Bench~2.}
\label{tab:transfer-details}
\centering
\small
\begin{tabular}{lllcc}
\toprule
Benchmark & Harness & Variant & $\mathrm{Pass}@8$ & $\mathrm{Avg}@8$ \\
\midrule
\multirow{10}{*}{$\tau^2$-bench} & \multirow{5}{*}{MiniSWE} & Seed & 96.2 & 84.0 \\
& & AHE & 95.6 & 86.7 \\
& & HyperAgents & 95.9 & 84.8 \\
& & A-Evolve & 96.0 & 85.4 \\
& & GEPA & 97.8 & 86.6 \\
\cmidrule(l){2-5}
& \multirow{5}{*}{Codex} & Seed & 97.5 & 87.0 \\
& & AHE & 96.8 & 85.3 \\
& & HyperAgents & 96.8 & 87.3 \\
& & A-Evolve & 96.5 & 85.0 \\
& & GEPA & 96.5 & 85.6 \\
\midrule
\multirow{10}{*}{DeepSWE~v1.1} & \multirow{5}{*}{MiniSWE} & Seed & 57.5 & 23.5 \\
& & AHE & 59.3 & 21.1 \\
& & HyperAgents & 61.1 & 21.3 \\
& & A-Evolve & 61.9 & 23.2 \\
& & GEPA & 54.9 & 21.8 \\
\cmidrule(l){2-5}
& \multirow{5}{*}{Codex} & Seed & 77.0 & 45.1 \\
& & AHE & 85.0 & 45.0 \\
& & HyperAgents & 83.2 & 48.2 \\
& & A-Evolve & 78.8 & 43.0 \\
& & GEPA & 80.5 & 46.6 \\
\bottomrule
\end{tabular}
\par
\end{table}

\section{Mechanisms: Additional Protocols and Analyses}
\label{app:rq3-details}

This appendix supplements the Mechanisms analysis in Section~\ref{sec:mechanism} with Parent expansion, Component comparisons, and admission replay protocols. Appendix~\ref{app:component-definitions} details Component implementations and case selection.

\subsection{Select: Parent Choice and Subsequent Improvement}
\label{app:rq3-select-details}

\textbf{Parent Pools and Common Generation Process.} For each benchmark and harness, we form one pool containing the frozen Champions from AHE, HyperAgents, A-Evolve, and GEPA. The four pools contain 16 Parents. Each Parent initializes three separate branches from the same frozen workspace, producing 48 Candidates in total. Every branch uses the same original A-Evolve \emph{analyze} and \emph{mutate} implementations, prompts, and generation allowance. The four methods provide Parent artifacts; the shared downstream process holds the modification procedure fixed.

\textbf{Execution Evidence and Evaluation.} Each Parent executes the 50 Train tasks once, and its three branches share this evidence while running analysis and mutation independently. Canonical Validation uses one pass over 20 tasks on $\tau^3$-Banking or 19 on Terminal-Bench~2. Each Parent and Candidate receives eight Test passes over 27 or 20 tasks, respectively. Target and mutation models and harness versions follow Section~\ref{sec:analysis-procedure}. Test outcomes determine the reporting scores and best-Parent labels. Candidate scores enter this analysis regardless of the original A-Evolve admission verdict, so the diagnostic measures the changes produced from each Parent.

\textbf{Original Selection Rules.} We apply the frozen HyperAgents, A-Evolve, and GEPA \emph{select} rules to each common four-Parent pool. HyperAgents uses its native score-weighted sampling rule with Train evidence. A-Evolve uses the one-pass Validation aggregate, and GEPA uses the corresponding per-task Validation vector. Diagnostic metadata, child-count initialization, and tie handling are fixed before these rules are evaluated. AHE's chronological selector requires a shared lineage order, so AHE contributes a Parent artifact and the other three rules define Original. Neither Test outcomes nor sibling Candidate results enter these selection decisions or the analysis and mutation inputs.

\Needspace{9\baselineskip}
\textbf{Best Parent and Hit Rate.} Let $\mathcal{P}$ be a four-Parent pool, $C_{P,j}$ the Candidate from branch $j$ of Parent $P$, and $S(A)$ the Test $\mathrm{Avg}@8$ of agent $A$. We define the mean improvement and the best-Parent set as
\begin{equation}
\Delta(P)=\frac{1}{3}\sum_{j=1}^{3}S(C_{P,j})-S(P),
\qquad
\mathcal{B}=\arg\max_{P\in\mathcal{P}}\Delta(P).
\label{eq:parent-improvement}
\end{equation}
For the three original rules $\mathcal{S}$, let $\pi_s(P)$ be the probability that rule $s$ selects $P$. Table~\ref{tab:select-parent-quality} reports
\begin{equation}
H_{\mathrm{Original}}=\frac{100}{3}\sum_{s\in\mathcal{S}}\sum_{P\in\mathcal{B}}\pi_s(P),
\qquad
H_{\mathrm{Random}}=100\frac{|\mathcal{B}|}{4}.
\label{eq:parent-hit-rate}
\end{equation}
A deterministic rule contributes zero or one before aggregation; a stochastic rule contributes its probability of selecting a best Parent. Each pool has a unique maximizer, giving a 25\% random baseline. Improvement remains signed, so the best Parent is the one with the largest gain or smallest decline.

\textbf{Current Performance and Improvement Potential.} In all four pools, the Parent with the highest Test score differs from the Parent with the largest mean improvement. On general tasks, the best-Parent gain is $-2.0$ points for the source-editable harness and $+19.6$ for the outer-layer-only harness; on coding tasks, it is $+4.0$ and $+1.5$, respectively. The original rules' mean hit rate falls below random on general tasks and exceeds it on coding tasks for both harnesses. These comparisons motivate assessing selection by subsequent improvement under a common generation process.

\subsection{Analyze: Component and Reminder Comparisons}
\label{app:rq3-analyze-details}

\textbf{Comparison Logic.} Applying a Component to the matching Seed measures its standalone effect; comparing that variant with Champion tests how the remaining changes affect its benefit. Generic-reminder controls test whether broad guidance reproduces those effects. These comparisons inform \emph{analyze} through the specificity of useful guidance and \emph{mutate} through the composition of changes.

\textbf{A Recurring Specificity Advantage.} Fifteen of sixteen Component variants have higher reliability than their matching generic-reminder control, and every Champion case has at least one such Component. The exception is the isolated shell-adaptation Component in the reflection-centric, source-editable coding-task case, at 64.0 versus 65.9 for the reminder; its task-guidance Component instead reaches 67.3. Useful specific changes are therefore not reproduced by general advice, even though not every isolated Component helps. Reminders leave source-editable reliability close to Seed on both task domains and lower outer-layer-only reliability on both.

\textbf{Case Composition.} The eight Component cases cover both harness types and task domains. Seven are implementation-centric; the reflection-centric case uses the source-editable harness on coding tasks. Each case pairs concrete Champion changes with a generic-reminder intervention on the matching Seed.

\textbf{Diagnosis Requirements.} The specificity advantage motivates a diagnosis contract that links failure explanations to execution traces, identifies useful behavior to preserve, and states when the proposed correction applies. We assess this contract through the changes and outcomes it produces.

\subsection{Mutate: Component Effects and Interactions}
\label{app:rq3-mutate-details}

\textbf{Complete Revisions Usually Outperform Subsets.} Six of eight Champions have higher reliability than either isolated Component, including two implementation-centric general-task cases and all four coding-task cases. The exceptions are two other implementation-centric general-task cases, one per harness type: the source-editable interface Component reaches 34.4 versus 32.6 for Champion, and the outer-layer-only orchestration Component reaches 37.2 versus 36.9. In the latter case, skills alone also increase coverage to 72.2 versus 66.0, despite slightly lower reliability. Conversely, an implementation-centric coding-task Champion with the outer-layer-only harness reaches 72.6 reliability even though both isolated variants, at 69.1 and 68.0, fall below Seed's 69.9. These results motivate evaluating combinations alongside isolated effects.

\textbf{Interaction Definition and Estimates.} For the four completed $\tau^3$-Banking decompositions, let $S_0$, $S_A$, $S_B$, and $S_{AB}$ denote $\mathrm{Avg}@8$ for Seed, Seed + A, Seed + B, and Champion. We measure non-additivity by
\[
I=(S_{AB}-S_0)-[(S_A-S_0)+(S_B-S_0)].
\]
A negative value means that the joint gain is smaller than the sum of the two standalone gains. The Champions from AHE have interactions of 0.0 points for MiniSWE (95\% task-paired interval $[-4.5,4.9]$) and $+2.1$ for Codex ($[-2.2,6.2]$), with both intervals including zero. The Champions from HyperAgents have negative interactions for MiniSWE ($-6.3$; $[-11.7,-1.2]$) and Codex ($-6.6$; $[-12.6,-0.6]$). These cross-harness observations identify interference as a design risk in the sampled implementation-centric general-task cases.

\textbf{Compatibility Checks.} Interaction estimates use the four verified $\tau^3$-Banking decompositions, while Terminal-Bench~2 provides targeted subset comparisons. Negative interactions motivate checking whether a Candidate diff follows its diagnosis, includes necessary dependencies, and preserves useful behavior. Coupled changes remain permissible when their dependencies and rationale are explicit.

\subsection{Gate: Size-Normalized Best-Candidate Retention}
\label{app:rq3-gate-details}

\textbf{Candidate Pools.} For each benchmark, harness, and method, we freeze four distinct artifacts from the same evolution lineage: the Seed, historical Champion, last legal Candidate, and Candidate nearest 50\% of cumulative recorded evolution cost. Predeclared 25\% and 75\% checkpoints resolve duplicates. This gives eight pools per benchmark. Each artifact has eight-pass measurements on separate decision and reporting panels of 20 and 27 tasks for $\tau^3$-Banking, or 19 and 20 for Terminal-Bench~2. These task panels follow the canonical diagnostic splits.

\textbf{Original Admission Policies.} Original gate uses the initial implementations in Table~\ref{tab:method-operators}. AHE, HyperAgents, and GEPA accept all eligible Candidates with completed evaluations; A-Evolve requires strict mean improvement over its matched Parent. The comparison supplies four legal, fully evaluated Candidates to these fixed admission policies, with upstream proposal checks held outside the replay.

\textbf{Rules and Replay Contract.} Each pool starts with Seed as incumbent and processes the historical Champion, last legal Candidate, and cost-indexed checkpoint in the frozen order. This is a diagnostic order, not the historical generation order. An accepted challenger becomes the incumbent for subsequent comparisons, while all previously retained Candidates remain in the set. The constrained rule is unchanged from the original replay: it requires strict mean improvement over the incumbent on the same purchased evidence, with ties rejected, and
\begin{equation}
L=\frac{1}{|\mathcal{D}_{g}|}\sum_{x\in\mathcal{D}_{g}}\max\{0,\hat p_{\mathrm{inc}}(x)-\hat p_{\mathrm{cand}}(x)\}\leq 0.05,
\label{eq:gate-regression}
\end{equation}
where $\mathcal{D}_{g}$ contains 20 decision tasks on $\tau^3$-Banking or 19 on Terminal-Bench~2, and each success fraction uses passes 1--2. All policies receive the same 160 or 152 unique Candidate--task trials, respectively; eligibility-only gates need not use their numeric scores. For A-Evolve's original gate, the matched parent is the current replay incumbent. A separate reporting panel of 27 or 20 tasks, respectively, uses eight passes to identify the empirical best and never supplies decision evidence. Both benchmarks use the same fixed order, strict-improvement rule, regression cap, and retained-set definition.

\textbf{Metric Definitions.} For method $m$ and gate $g$, let $\mathcal{P}_m$ be its four-Candidate input pool, $\mathcal{R}_{m,g}$ the retained set including Seed, and $\mathcal{B}_m=\arg\max_{c\in\mathcal{P}_m}\mathrm{Avg}@8(c)$ on the reporting panel. The hit indicator is $H_{m,g}=\mathbf{1}[\mathcal{R}_{m,g}\cap\mathcal{B}_m\ne\varnothing]$ and set size is $N_{m,g}=|\mathcal{R}_{m,g}|$. Retaining any tied maximizer counts as a hit. The size-normalized score is $H_{m,g}/N_{m,g}$; Seed ensures a nonempty retained set. Each main-table cell reports $100\sum_m(H_{m,g}/N_{m,g})/4$, separately for each benchmark and harness. This is the mean of per-pool ratios, not the ratio of mean hit to mean size. Values are rounded to one decimal place only after aggregation. All sixteen pools have a unique empirical best. The score therefore equals the chance of drawing that Candidate uniformly from the retained set; this interpretation does not prescribe the method's actual \emph{select} policy.

\begin{table}[htbp]
\caption{Per-pool audit of size-normalized best-Candidate retention. Each policy cell gives $H/N$, the binary best-retention indicator divided by set size, including Seed. The empirical best maximizes reporting-panel $\mathrm{Avg}@8$; Generation numbers identify the frozen artifacts.}
\label{tab:gate-retention-cases}
\centering
\small
\setlength{\tabcolsep}{4pt}
\begin{tabularx}{\textwidth}{@{}lllYcc@{}}
\toprule
\multirow{2}{*}{Benchmark} & \multirow{2}{*}{Harness} & \multirow{2}{*}{Method} & \multirow{2}{*}{Empirical best} & Original gate & Constrained \\
& & & & $H/N$ & $H/N$ \\
\midrule
\multirow{8}{*}{$\tau^3$-Banking} & \multirow{4}{*}{MiniSWE} & AHE & Generation 9 & $1/4$ & $0/1$ \\
 &  & HyperAgents & Generation 3 & $1/4$ & $1/2$ \\
 &  & A-Evolve & Generation 1 & $1/2$ & $1/2$ \\
 &  & GEPA & Generation 9 & $1/4$ & $0/1$ \\
\cmidrule(lr){2-6}
 & \multirow{4}{*}{Codex} & AHE & Generation 10 & $1/4$ & $0/2$ \\
 &  & HyperAgents & Generation 11 & $1/4$ & $0/2$ \\
 &  & A-Evolve & Generation 20 & $0/2$ & $0/2$ \\
 &  & GEPA & Seed & $1/4$ & $1/1$ \\
\midrule
\multirow{8}{*}{Terminal-Bench~2} & \multirow{4}{*}{MiniSWE} & AHE & Generation 18 & $1/4$ & $0/1$ \\
 & & HyperAgents & Generation 10 & $1/4$ & $1/2$ \\
 & & A-Evolve & Generation 4 & $1/4$ & $0/1$ \\
 & & GEPA & Generation 20 & $1/4$ & $0/2$ \\
\cmidrule(lr){2-6}
 & \multirow{4}{*}{Codex} & AHE & Generation 20 & $1/4$ & $0/1$ \\
 & & HyperAgents & Generation 18 & $1/4$ & $0/1$ \\
 & & A-Evolve & Seed & $1/1$ & $1/1$ \\
 & & GEPA & Seed & $1/4$ & $1/1$ \\
\bottomrule
\end{tabularx}
\par
\end{table}

\textbf{A Recurring Preservation Risk.} The original policies retain the reporting-panel best in fifteen pools, compared with six under Constrained. The nine newly excluded best Candidates span both harness types, task domains, and paradigms. Size-normalized scores nevertheless improve in four pools, remain unchanged in three, and fall in nine. The improvements include two implementation-centric source-editable pools retaining the best among fewer alternatives, and two reflection-centric outer-layer-only pools retaining a best Seed alone. Pruning can therefore improve the diagnostic metric while excluding useful Candidates elsewhere. The two outer-layer-only averages increase even while evolved best Candidates are lost in other pools. Jointly reporting hit, set size, and Candidate identity distinguishes concentration on a strong fallback from preservation of evolved gains.

\textbf{Replay Evidence and Design Implications.} The replay applies original and constrained admission rules to the same frozen Candidate pools and measured scores. The constrained rule combines strict score improvement with a regression cap. Comparing retained-set quality, size, and membership motivates admission rules that balance improvement, preservation, and fallback retention.

\subsection{Component Definitions and Representative Content}
\label{app:component-definitions}

This subsection records the frozen Component definitions for the eight-case matrix of Table~\ref{tab:mechanism-factorization}. All sixteen Component variants and eight Champion comparators have complete evaluations. A and B identify the two groups within each case. The definitions specify retained agent changes, implementation files, execution dependencies, and representative content. Prompt and code excerpts have line wrapping adjusted for display, and code excerpts reproduce the indicated implementation segment. MiniSWE paths are relative to \path{src/minisweagent/}; Codex paths are relative to the Candidate's target directory. Each Component is evaluated as a complete agent intervention.

\textbf{Construction and Comparison.} Seed + A and Seed + B retain the matching Seed outside the specified Component and its execution dependencies. Components are separated by function: a configuration file can contain both interface settings and task-solving instructions, which we isolate by semantic key. Required compatibility adaptations are identified below. Codex Components remain outside its immutable executable and source. The Terminal-Bench~2 cases evaluate targeted Champion subsets. Interaction estimates use decompositions whose joint Components reproduce the compared Champion.

\textbf{Case Selection.} The matrix contains four Champion cases per benchmark, with two from each harness. Eligibility requires at least two functionally distinct groups of persistent changes that can be isolated with their execution dependencies on the matching Seed while keeping the model and evaluator fixed. The $\tau^3$-Banking cases comprise Champions from AHE and HyperAgents in both harnesses. On Terminal-Bench~2, we retain the previously evaluated Champions from MiniSWE with AHE and Codex with HyperAgents, and add MiniSWE with GEPA and Codex with AHE after auditing all eight Champions. The additions prioritize clear functional boundaries and complementary mechanisms. Component boundaries were frozen before evaluation, and all completed outcomes are retained.

\textbf{Selection History and Extension.} The initial Terminal-Bench~2 sample selected the largest historical unrounded $\mathrm{Avg}@8$ gain within each harness, yielding MiniSWE with AHE and Codex with HyperAgents. Case identities remain frozen, and source-faithful reevaluations replace the two superseded Codex Champion measurements. MiniSWE with GEPA separates task guidance from shell execution adaptation, complementing AHE's broader validation and submission controls. Codex with AHE separates instruction content from persistent context and skill support, complementing HyperAgents' dependency bootstrap. Functional clarity and mechanism complementarity determine the additions. Their identities and Component boundaries were frozen before evaluation, and all added outcomes are reported.

\subsubsection{\texorpdfstring{$\tau^3$}{Tau3}-Banking: MiniSWE with AHE (G2)}
\label{app:components-tau3-miniswe-ahe}

\textbf{Component A: Interface.} This Component changes how shell/tool output is presented to the model, while retaining the Seed's task-solving instructions and bash action interface. It extracts readable text from structured tool-response envelopes, removes duplicate text blocks, and preserves salient information when an observation is long. Rather than retaining only its beginning and end, the observation also includes a bounded selection of middle content containing tool schemas, execution results, required fields, and conversation-control signals. The changed files are \path{models/utils/actions_toolcall.py}, \path{models/utils/actions_toolcall_response.py}, the added \path{models/utils/observation_output.py}, and the \texttt{observation\_template} key in \path{config/mini.yaml}.

\begin{quote}
\small
\textit{Observation-budget constants from the added helper:}
\begin{verbatim}
DISPLAY_OUTPUT_LIMIT = 10000
DISPLAY_OUTPUT_HEAD_CHARS = 3000
DISPLAY_OUTPUT_TAIL_CHARS = 3000
SALIENT_OUTPUT_MAX_CHARS = 2500
\end{verbatim}
\end{quote}

\textbf{Component B: Strategy.} This Component replaces the system prompt, task prompt, and format-error guidance in \path{config/mini.yaml}; observation processing stays at the Seed. The instructions distinguish the benchmark's tool-driven general tasks from coding tasks, prioritize live state and tool schemas, require one focused action followed by inspection, prohibit invented arguments or case facts, and distinguish a stopped conversation from a successfully completed backend action. The format-error guidance also specifies concise recovery from token-limit truncation and a tool-only final submission.

\begin{quote}
\small
\textit{Task-prompt excerpt:} ``Prefer live environment state, exact tool outputs, runtime status, and tool schemas over inference from general KB text.''

\textit{Completion-rule excerpt:} ``Only treat the task as complete when the required environment state or required action has actually been achieved.''
\end{quote}

\subsubsection{\texorpdfstring{$\tau^3$}{Tau3}-Banking: MiniSWE with HyperAgents (G9)}
\label{app:components-tau3-miniswe-hyperagents}

\textbf{Component A: Interface.} This Component exposes structured benchmark tools in place of a bash-only interface when it detects tasks from the general-task benchmark. It adds tool schemas, argument parsing (including nested JSON-string normalization), dispatch to the benchmark environment, and a dedicated final-submission action. The dependency-closed implementation spans \path{agents/interactive.py}, \path{environments/local.py}, \path{utils/tau3.py}, both action/tool-response utilities, and seven adapters in \path{models/}: the chat and Responses implementations for LiteLLM, OpenRouter, and Portkey, plus Requesty. The thirteenth changed file, \path{config/mini.yaml}, adds neutral transport and format-error instructions so that the Seed prompt does not request an unavailable bash tool. This compatibility text describes how to call the exposed tools, without importing the Champion's task-solving strategy.

\begin{quote}
\small
\textit{Exposed tool names:} \path{tau3_call}, \path{tau3_list_tools}, \path{tau3_get_tool_schema}, and \path{submit_final_output}.

\textit{Compatibility-note excerpt:} ``This note describes only the available transport; determine the task policy and actions from the task, tool schemas, and runtime responses.''
\end{quote}

\textbf{Component B: Strategy.} This Component changes only the task-instance prompt in \path{config/mini.yaml}. It retains the Champion's service workflow: start the conversation once, consult the latest user message and schema, perform one action at a time, separate user communication from domain-tool calls, respect stop/transfer signals, and ground advice in documented policy and hard user requirements. Because the Seed does not expose Component A's structured tools, the workflow is expressed through the Seed's \texttt{tau3-mcp.py} command-line interface. This is the evaluated CLI-translated strategy Component, not an unmodified copy of the native-tool prompt; the translation is an execution dependency, not an independent strategy finding.

\begin{quote}
\small
\textit{Translated task-prompt excerpts:} ``Perform exactly one focused Tau3 CLI action per response and inspect its output before the next action.'' ``Do not recommend a product or action that misses a user-stated hard requirement.''
\end{quote}

\subsubsection{\texorpdfstring{$\tau^3$}{Tau3}-Banking: Codex with AHE (G15)}
\label{app:components-tau3-codex-ahe}

\textbf{Component A: Orchestration.} The only changed file is the outer adapter \path{agent.py}. Before invoking Codex, it detects service-task markers in the instruction and appends a workflow-guidance block, unless that block is already present. The injected text emphasizes exact tool discovery, a per-entity worklist, assistant-side versus user-side actions, date and eligibility checks, canonical write arguments, and completing all requested items before ending. The Seed's prompt and skill files are retained. Here, \emph{orchestration} includes conditional instruction delivery; it does not mean that the Component contains no natural-language guidance.

\begin{quote}
\small
\textit{Injected-guidance excerpt:} ``Keep a live per-entity worklist for the case. For each named transaction, card, account, product, referral, or other requested item, track the exact entity, the user-described problem, the currently grounded workflow or enum, any missing fields, and the next read or write needed.''
\end{quote}

\textbf{Component B: Prompt and Skills.} This Component replaces \path{prompt.md} and \path{skills/task-execution/SKILL.md}, and adds \path{skills/tau3-service-workflows/SKILL.md}, while retaining the Seed adapter. The prompt distinguishes coding work from the general-task benchmark's workflows and directs the latter to the specialized skill. That skill specifies a six-part workflow: discover the live tool surface; ground on the exact entity; identify who must perform the next state change; construct exact writes; preserve the selected product or workflow branch; and finish all requested work. It includes explicit checks for relative dates, enums, optional fields, user-side follow-through, and multi-item completion.

\begin{quote}
\small
\textit{Specialized-skill excerpt:} ``Once the user accepts a supported product or workflow branch, stop re-optimizing and execute that exact branch.''

\textit{Prompt excerpt:} ``If the task involves multiple requested actions or entities, keep a short checklist and do not finish while any requested item is unresolved.''
\end{quote}

\subsubsection{\texorpdfstring{$\tau^3$}{Tau3}-Banking: Codex with HyperAgents (G18)}
\label{app:components-tau3-codex-hyperagents}

\textbf{Component A: Orchestration.} The outer \path{agent.py} detects the configured service server or service markers in the instruction, then prepends a guidance block unless its header is already present. The same server-detection helper is reused by the existing installation path. The injected guidance prioritizes live account investigation, eligible and actionable recommendations, exact knowledge-base procedures, discoverable-tool checks, literal argument formatting, and completion of accepted workflows. Prompt and skill files remain at the Seed, so the specialized-skill reference in this block does not itself install Component B.

\begin{quote}
\small
\textit{Injected-guidance excerpt:} ``Do not substitute a neighboring tool or an incidental side effect for a separately documented step. If the KB names a dedicated replacement, approval, closure, report, or submission tool, use that exact tool instead of assuming another mutation already covered it.''
\end{quote}

\textbf{Component B: Skills.} This Component changes \path{skills/task-execution/SKILL.md} to route tasks from the general-task benchmark to the added \path{skills/tau3-banking-runtime/SKILL.md}; the outer adapter and main prompt stay at the Seed. The specialized skill covers identity verification, account-specific lookups, recommendation eligibility, tool unlocking and execution checks, multi-step discrepancies, transfers, closures, and final-state verification. It also includes implementation-sensitive payload rules: preserve surfaced key order, whitespace, explicit null/false values, and numeric forms in nested argument strings. These are actual retained instructions, not a claim that equivalent JSON representations generally have different semantics.

\begin{quote}
\small
\textit{Payload-formatting excerpt:} ``Serialize that inner JSON as a single-line object with double-quoted keys, the surfaced key order, and standard spaces after each comma and colon.''

\textit{Tool-check excerpt:} ``After \texttt{unlock\_discoverable\_agent\_tool}, check that the returned tool name, description, and parameters match the tool you meant to unlock.''
\end{quote}

\subsubsection{Terminal-Bench~2: MiniSWE with AHE (G18)}
\label{app:components-terminal-miniswe-ahe}

\textbf{Component A: Static Instructions.} Only \path{config/mini.yaml} changes. The retained prompt additions call for early inspection of visible tests and checkers, exact deliverable paths and formats, fresh-shell verification, preservation of original inputs and consumers, and rechecking after any subsequent change. They reject proxy checks, fabricated substitutes, and unresolved failures as evidence of completion. These instructions do not add a programmatic submission guard.

\begin{quote}
\small
\textit{Contract-first instruction:} ``Before long implementation, downloads, or external research, spend the first 1-2 commands locating the real acceptance path.''

\textit{Reverification instruction:} ``Any edit after a passing verification invalidates that verification. If you change the artifact, environment, dependency set, or generated output afterward, rerun the exact end-to-end check against the final state before submitting.''
\end{quote}

\textbf{Component B: Dynamic Guard.} Files \path{agents/default.py} and \path{environments/local.py} implement the dynamic checks. The configuration file \path{config/mini.yaml} supplies their hooks and activation flags, without Component A's static instructions. It discovers visible verifier files, derives deliverable-boundary hints from the task, supplies fresh-shell compatibility helpers, records test inspection and unresolved check failures, and inspects submission attempts. The guard can replace a completion signal with a warning about uninspected verifiers, unhealthy package state, unresolved failures, missing files, or deliverable-boundary violations. It warns once per warning fingerprint; repeating the same submission can proceed, so this is not an unconditional veto or a replacement for the benchmark evaluator.

\begin{quote}
\small
\textit{Activated flags in the configuration:}
\begin{verbatim}
pause_submit_for_uninspected_verifier: true
pause_submit_for_unhealthy_system_packages: true
pause_submit_for_unresolved_verifier_failures: true
pause_submit_for_deliverable_boundary_violations: true
\end{verbatim}
\end{quote}

\subsubsection{Terminal-Bench~2: MiniSWE with GEPA (G20)}
\label{app:components-terminal-miniswe-additional}

\textbf{Component A: Task Guidance.} This Component combines static instructions about exact deliverables and evidence preservation with task-conditioned reminders. In \path{agents/default.py}, \path{_build_task_guardrails} detects wording associated with exhaustive outputs or fragile evidence and generates the corresponding guidance. The template in \path{config/mini.yaml} consumes the generated \path{task_guardrails} value and includes the accompanying static instructions. The generator and template hook stay in the same Component because either alone disconnects the mechanism. The local command-execution implementation remains at the Seed.

\textbf{Component B: Shell Execution Adaptation.} Only \path{environments/local.py} changes. The helper \texttt{\_preferred\_shell\_executable} resolves \texttt{MSWEA\_LOCAL\_SHELL}, defaulting to \texttt{bash}, and supplies the available executable to the subprocess launcher. It falls back to the platform default when the preferred shell cannot be resolved. This changes shell semantics at the command-execution boundary without importing Component A's guidance or changing the model configuration.

\subsubsection{Terminal-Bench~2: Codex with AHE (G20)}
\label{app:components-terminal-codex-additional}

\textbf{Component A: Prompt and Skills.} This Component replaces the prompt and task-execution skill in \path{prompt.md} and \path{skills/task-execution/SKILL.md}. The instructions prioritize the verifier-facing contract, a minimal coherent change on the active execution path, and checks of the final artifact at the consumer-visible boundary. Prompt and skill content form one guidance Component rather than separate Components based on file location. The adapter, compaction settings, model, reasoning effort, and Codex executable remain at the Seed.

\textbf{Component B: Persistent Context and Skill Support.} The outer \path{agent.py} moves temporary Codex state and skill-upload paths to a persistent Codex directory, while \path{codex.toml} enables prefix-preserving compaction with \texttt{auto\_compact\_token\_limit\_scope} set to \texttt{body\_after\_prefix}. These changes support instruction continuity without importing Component A's prompt or skill contents. The evolved adapter changes are semantically applied to the exact evaluation Seed, preserving its authentication, provider routing, session handling, and cancellation cleanup. Codex's executable and source, the model, and reasoning effort are unchanged.

\subsubsection{Terminal-Bench~2: Codex with HyperAgents (G18)}
\label{app:components-terminal-codex-hyperagents}

\textbf{Component A: Instructions.} The prompt and task-execution skill are replaced in \path{prompt.md} and \path{skills/task-execution/SKILL.md}; the outer adapter remains at the matching Seed. The prompt and skill instruct the agent to preserve authentic input artifacts, inspect visible verifiers, reconstruct missing checker dependencies, react to failed checks, match exact output paths and final-state requirements, remove probe-only artifacts, and run a final check from a fresh shell or process. They also ask the agent to diagnose silent long-running commands instead of repeatedly polling without new evidence.

\begin{quote}
\small
\textit{Skill excerpt:} ``Recreate the verifier runtime before declaring the task unavailable.''

\textit{Failure-handling excerpt:} ``Clear red checks before closing. If any command, test, assertion, or custom validation failed after your last meaningful edit, rerun that same path cleanly or keep debugging.''
\end{quote}

\textbf{Component B: Orchestration and Bootstrap.} Only the outer \path{agent.py} changes. The Champion's bootstrap changes are merged onto the exact Seed adapter used in the matched evaluation, rather than replacing that adapter with a different baseline wrapper. During setup, the adapter installs or locates \texttt{uv} and \texttt{uvx}, creates a sourceable environment helper, and exposes stable executable entrypoints in standard locations for fresh-shell verifier commands. Installation first attempts the Python package route and then an installer fallback; setup reports a warning if the bootstrap remains unavailable. The Seed's prompt and skills are retained. This Component prepares the execution environment; it does not alter the verifier's scoring rules or Codex's executable.

\begin{quote}
\small
\textit{Bootstrap function's documented purpose:} ``Install stable verifier entrypoints for uv-based test harnesses.''

\textit{Setup-hook excerpt:}
\begin{verbatim}
try:
    await self.exec_as_root(environment,
        command=_verifier_bootstrap_command())
except RuntimeError:
    self.logger.warning(
        "Failed to bootstrap uv verifier entrypoints",
        exc_info=True)
\end{verbatim}
\end{quote}

\textbf{Generic-Reminder Controls Are Separate.} The reminders in Table~\ref{tab:generic-reminder-controls} are not additional Champion Components. On $\tau^3$-Banking, both harnesses receive: ``This is an interactive tool-use task. Inspect the available runtime tool interface, use tools when needed, verify the resulting state, and complete every requested item before ending.'' The Terminal-Bench~2 MiniSWE control adds: ``Before final submission, inspect any available tests or check commands and run the closest exact end-to-end check on the final artifact. Do not submit while a relevant check is still failing or running.'' These controls add general advice without the case-specific prompt bundle, interface implementation, or dynamic guard above.

\textbf{Recovery of Interrupted Evaluations.} Four $\tau^3$-Banking Codex reminder evaluations were interrupted by API rate limits after their conversations had started. Under a documented recovery-policy amendment, we fixed these four task--pass identities and reran each once from its initial state with unchanged model, simulator, target, and evaluator settings. The other 772 numeric outcomes were retained, and all four replacement outcomes were accepted, including three zeros. Original attempts remain in the audit record; selected outcomes and their costs enter the reported metrics.

\section{Operator Redesign: Implementation and Evaluation}
\label{app:revision-protocols}

\subsection{Evidence and Implementation Requirements}

The analysis agent combines results from the Phenomenology, Generalization, and Mechanisms analyses with source traces and reference operators. It identifies recurring patterns across harnesses, task domains, and evolution paradigms and links each finding to the relevant comparisons and observed behavior. The resulting evidence report supplies the rationale and applicability conditions for operator revision.

Each proposed revision specifies the original rule, observed problem, implementation change, expected effect, and validation checks. The implementation agent receives this specification together with the evidence report, reference code, input/output contract, and permitted editing surface. It returns versioned code or instructions, evidence links, applicability conditions, fallback behavior, and validation records. Interface checks preserve the reference \emph{rollout}, \emph{validate}, and \emph{record} stages and each harness's editing contract.

\textbf{Diagnosis and Mutation Requirements.} The \emph{analyze} revision asks diagnoses to identify the observed failure and useful behavior to preserve. The \emph{mutate} revision asks Candidate changes to implement that diagnosis coherently and retain the required dependencies. These complementary contracts connect execution evidence to the resulting intervention.

\textbf{Revision Objectives.} The four revisions in Table~\ref{tab:design-hypotheses} target complementary decisions. Regression-aware \emph{select} uses preservation evidence to resolve score ties, behavior-grounded \emph{analyze} links diagnoses to execution traces, dependency-aware \emph{mutate} organizes coherent changes, and cost-aware \emph{gate} considers resource use at equal scores. These objectives guide implementation, expert review, and algorithm evaluation.

\subsection{Human Review Materials and Reporting}
\label{app:operator-human-review}

\begin{figure}[!htbp]
\centering
\includegraphics[width=\textwidth]{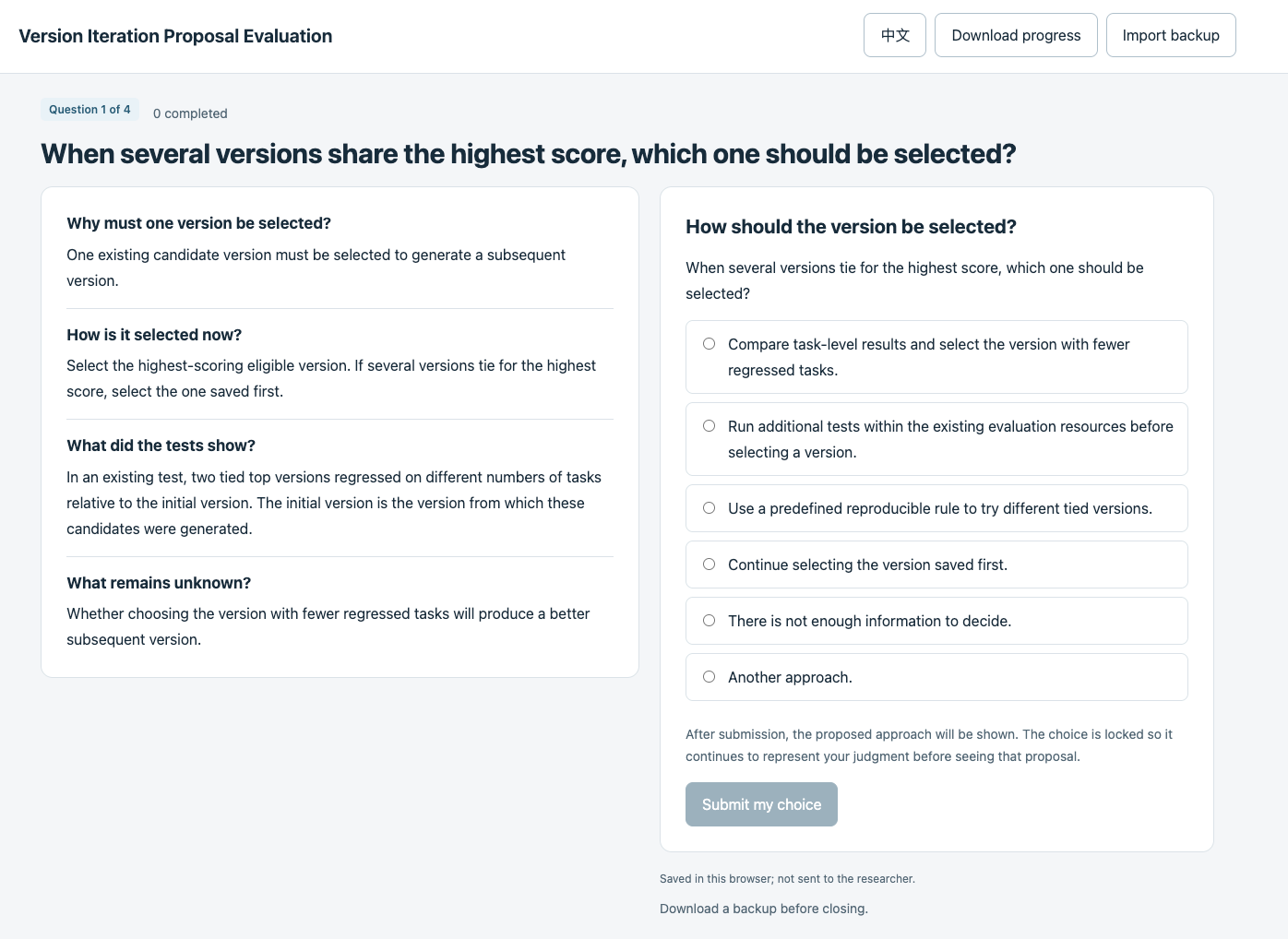}
\caption{Human review of \emph{select} before proposal disclosure. Experts inspect the original rule and evidence, then submit their preferred revision direction.}
\label{fig:human-review-direction}
\end{figure}

\begin{figure}[!htbp]
\centering
\includegraphics[width=\textwidth]{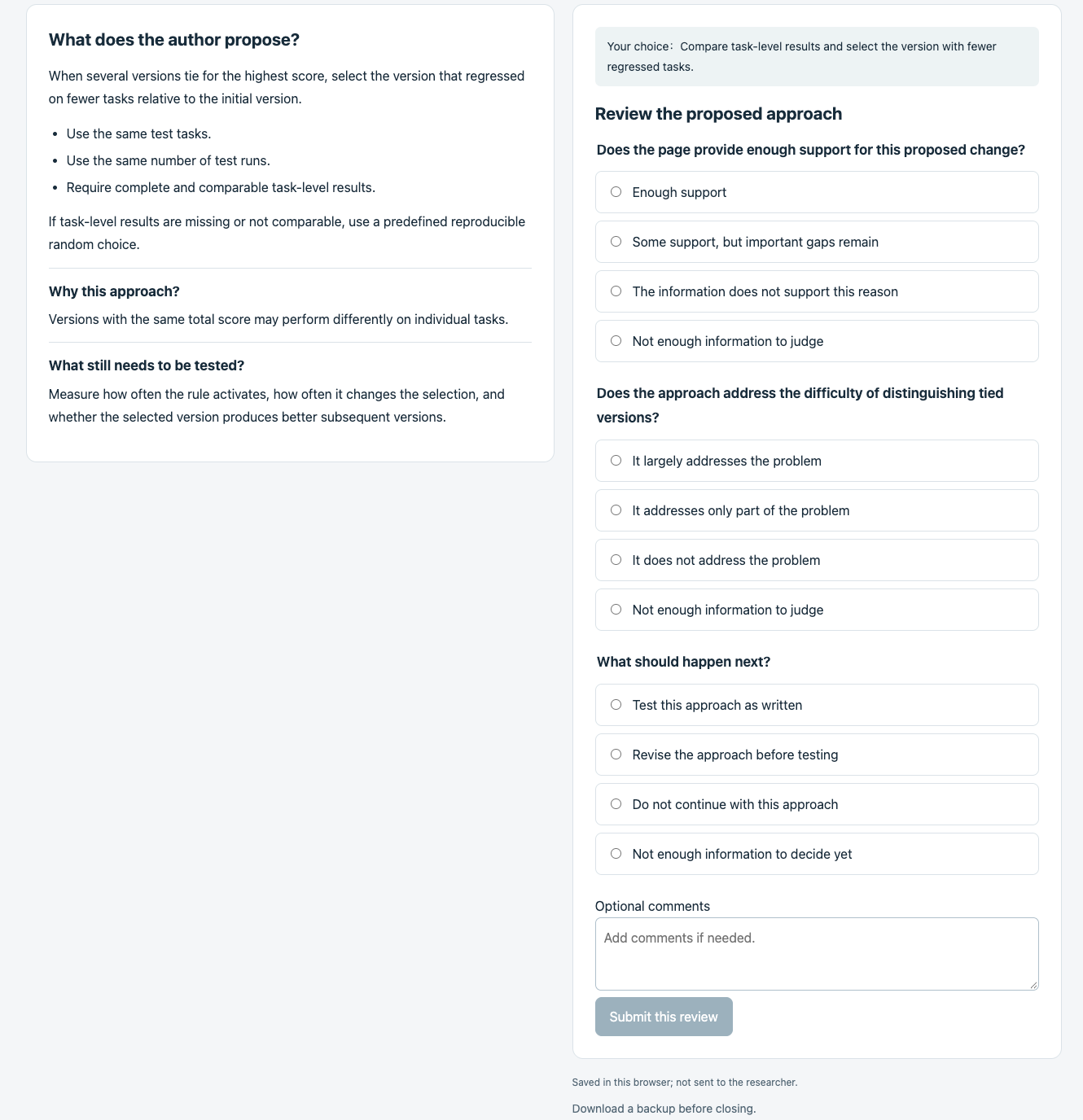}
\caption{Human review of \emph{select} after proposal disclosure. Experts inspect the revision and its reasoning, then assess evidence support, problem addressing, and endorsement for validation.}
\label{fig:human-review-proposal}
\end{figure}

Eight experts independently reviewed all four frozen operator revisions, with balanced case and direction-option orders. For each case, they inspected the original rule and evidence and submitted a direction before seeing EvoAudit's proposal. They then reviewed its reasoning, implementation changes, applicability, and validation plan, judging evidence support, problem addressing, and approval to validate. Each question included an insufficient-evidence option and optional comments. Figures~\ref{fig:human-review-direction} and~\ref{fig:human-review-proposal} show the direction-selection and proposal-review interfaces for the \emph{select} revision.

Table~\ref{tab:human-evaluation} reports mean positive counts per revision out of eight experts. Direction agreement uses a frozen mapping between the initial choice and EvoAudit's direction. Evidence support and problem addressing count fully positive responses, while endorsement counts approval to validate without further changes. Each measure sums positive counts across the four revisions and divides by four. Assignments, proposal versions, questions, and aggregation rules were frozen before formal review. All formal judgments are retained, and subsequent changes receive new proposal versions.

\subsection{Comparison with Existing Evolution Methods}

We compare archived 20-generation runs of AHE, HyperAgents, A-Evolve, and GEPA with separate 20-generation AGE runs on both primary benchmarks and harnesses. Each frozen Champion receives eight benchmark passes against its matching Seed. Table~\ref{tab:final-algorithm} reports reliability, coverage, task turnover, and deployment and evolution costs. Evidence-collection and analysis/design-agent costs are recorded separately from online evolution.

\FloatBarrier
\section{Evolution Trees}
\label{app:evolution-trees}

The following figures show all sixteen archived evolution runs, each comprising the Seed and 20 generation attempts. Unsuccessful attempts are retained. Arrows follow the recorded parent-to-child relationships, not chronological adjacency; layout positions and node colors do not indicate performance or acceptance decisions.

\begin{figure}[!htbp]
\centering
\includegraphics[width=\textwidth,height=0.70\textheight,keepaspectratio]{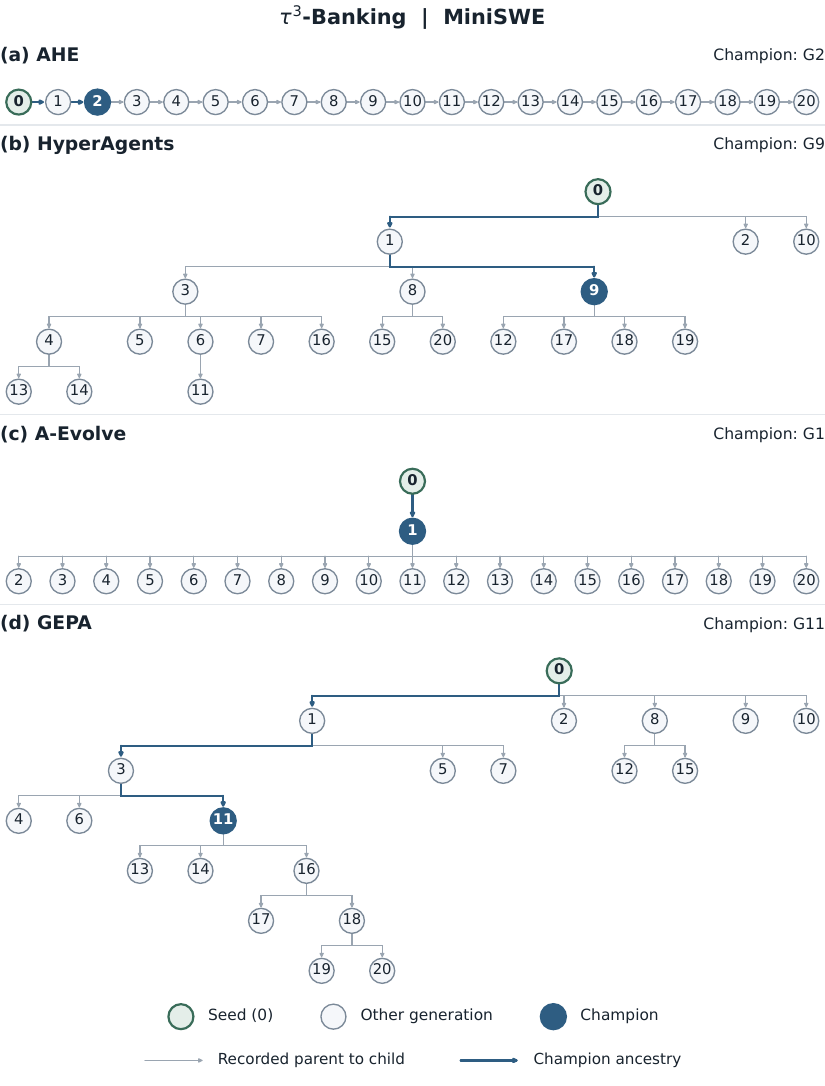}
\caption{Evolution trees on $\tau^3$-Banking with MiniSWE. Nodes are labeled by generation, with 0 denoting the Seed. Blue marks the Champion used in the paper and its ancestral path.}
\label{fig:evolution-trees-tau3-miniswe}
\end{figure}

\clearpage
\begin{figure}[htbp]
\centering
\includegraphics[width=\textwidth,height=0.86\textheight,keepaspectratio]{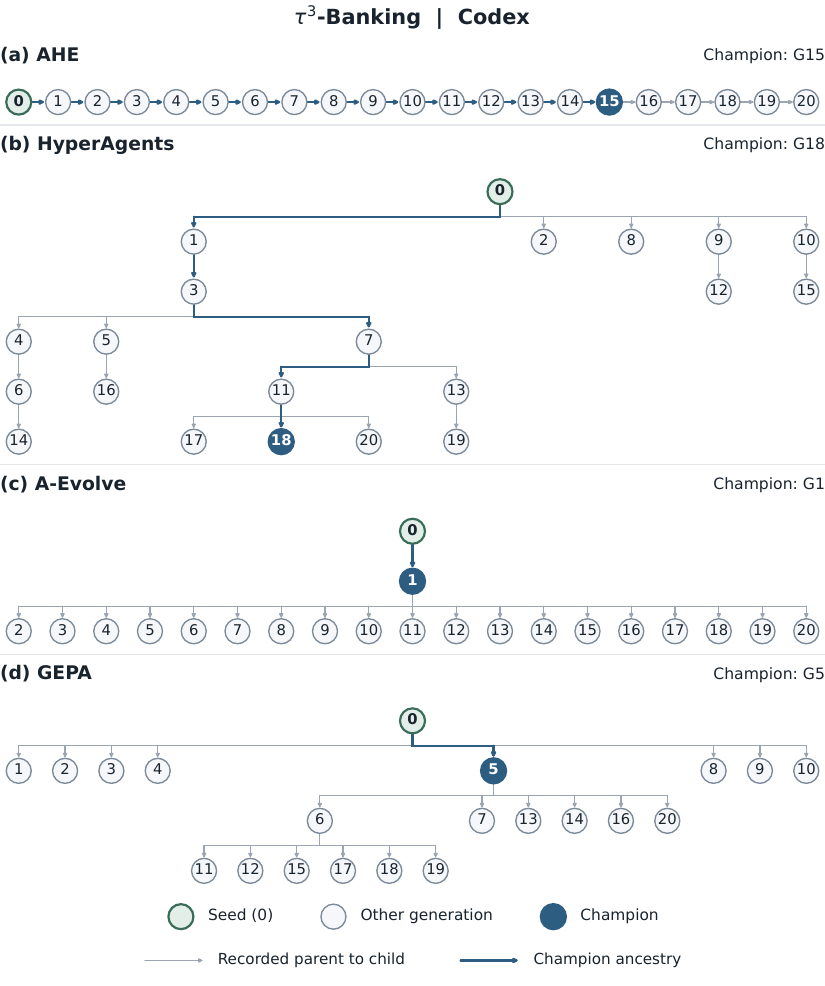}
\caption{Evolution trees on $\tau^3$-Banking with Codex. Nodes are labeled by generation, with 0 denoting the Seed. Blue marks the Champion used in the paper and its ancestral path.}
\label{fig:evolution-trees-tau3-codex}
\end{figure}

\clearpage
\begin{figure}[htbp]
\centering
\includegraphics[width=\textwidth,height=0.86\textheight,keepaspectratio]{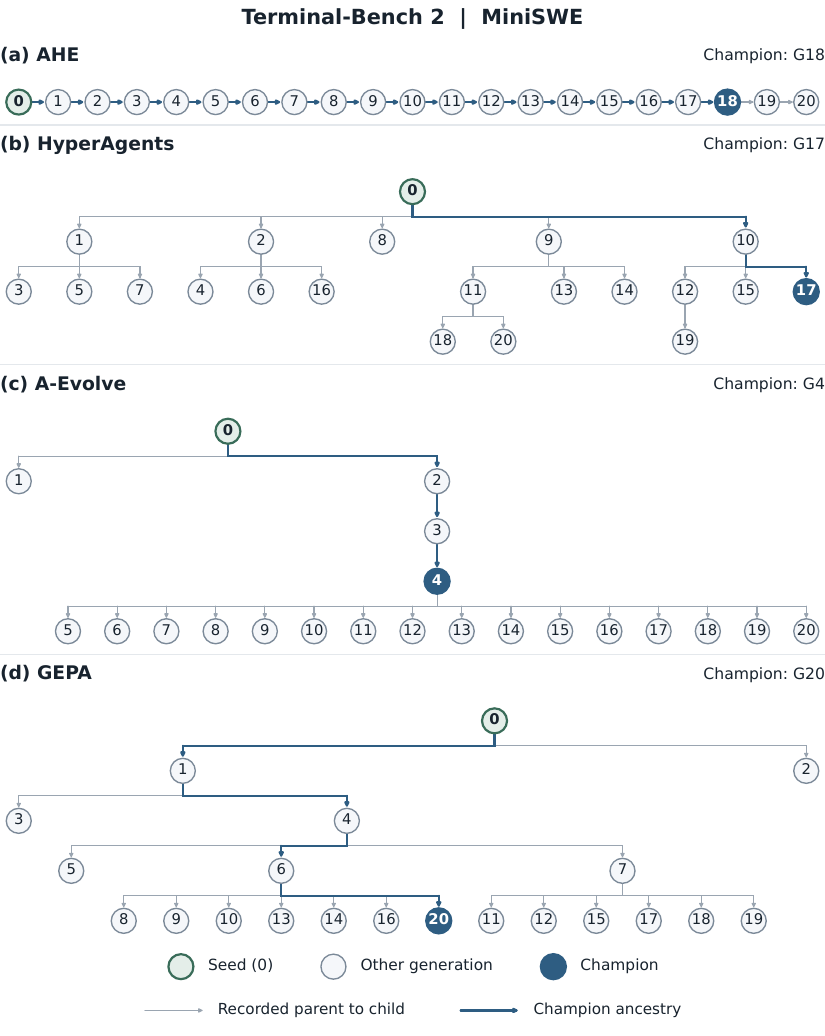}
\caption{Evolution trees on Terminal-Bench~2 with MiniSWE. Nodes are labeled by generation, with 0 denoting the Seed. Blue marks the Champion used in the paper and its ancestral path.}
\label{fig:evolution-trees-tb2-miniswe}
\end{figure}

\clearpage
\begin{figure}[htbp]
\centering
\includegraphics[width=\textwidth,height=0.86\textheight,keepaspectratio]{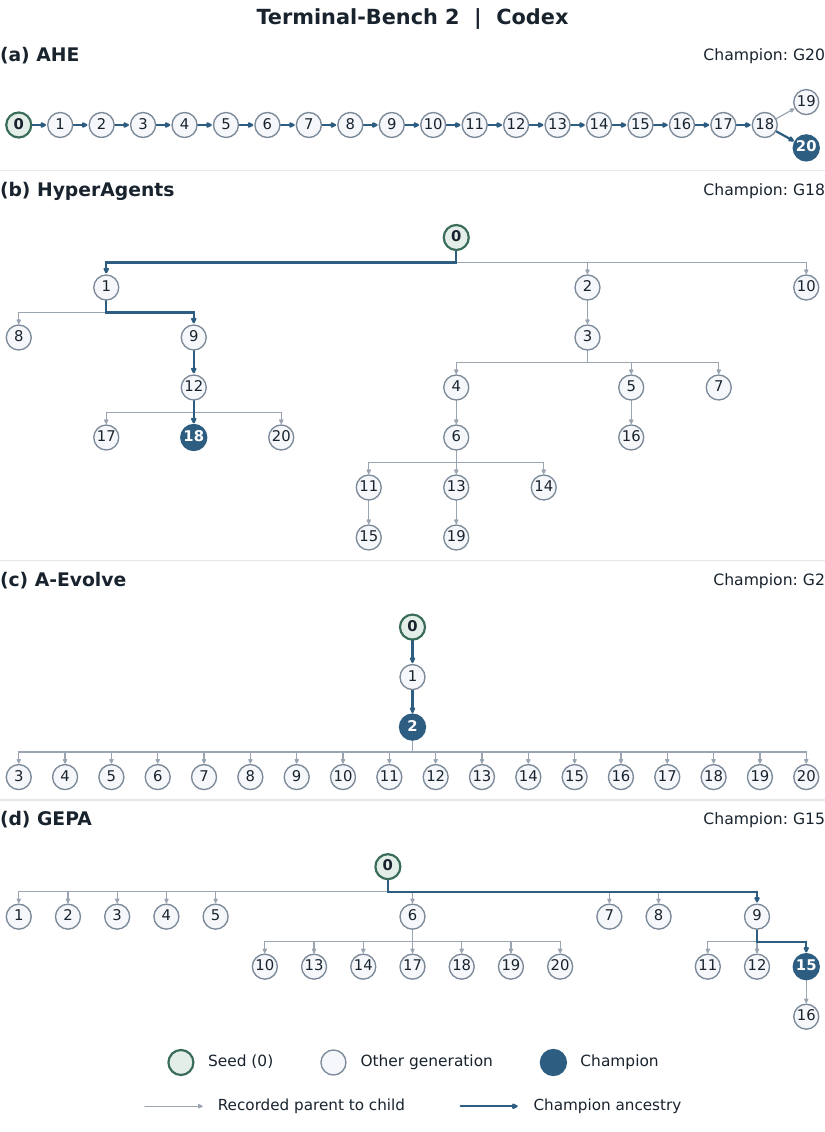}
\caption{Evolution trees on Terminal-Bench~2 with Codex. Nodes are labeled by generation, with 0 denoting the Seed. Blue marks the Champion used in the paper and its ancestral path.}
\label{fig:evolution-trees-tb2-codex}
\end{figure}

\end{document}